\documentclass[10pt,twocolumn,letterpaper]{article}

\usepackage[pagenumbers]{cvpr} % To force page numbers, e.g. for an arXiv version

\usepackage[dvipsnames]{xcolor}

\usepackage{amsmath}
\usepackage{amssymb}
\usepackage{adjustbox}
\definecolor{cvprblue}{rgb}{0.21,0.49,0.74}
\usepackage[pagebackref,breaklinks,colorlinks,citecolor=cvprblue]{hyperref}

\usepackage{multirow} 
\def\paperID{10299} % *** Enter the Paper ID here
\def\confName{CVPR}
\def\confYear{2024}

\begin{document}

\title{Spacetime Gaussian Feature Splatting for Real-Time Dynamic View Synthesis}

\author{Zhan Li$^{1,2*}$
\qquad
Zhang Chen$^{1\dagger}$
\qquad
Zhong Li$^{1\dagger}$
\qquad
Yi Xu$^{1}$\\
$^{1}$ OPPO US Research Center \qquad $^{2}$ Portland State University\\
{\tt\small lizhan@pdx.edu \qquad zhang.chen@oppo.com \qquad zhong.li@oppo.com \qquad yi.xu@oppo.com}\\
{\tt\small \url{https://oppo-us-research.github.io/SpacetimeGaussians-website/}}}

\makeatletter
\g@addto@macro\@maketitle{
  \includegraphics[width=1.0\textwidth]{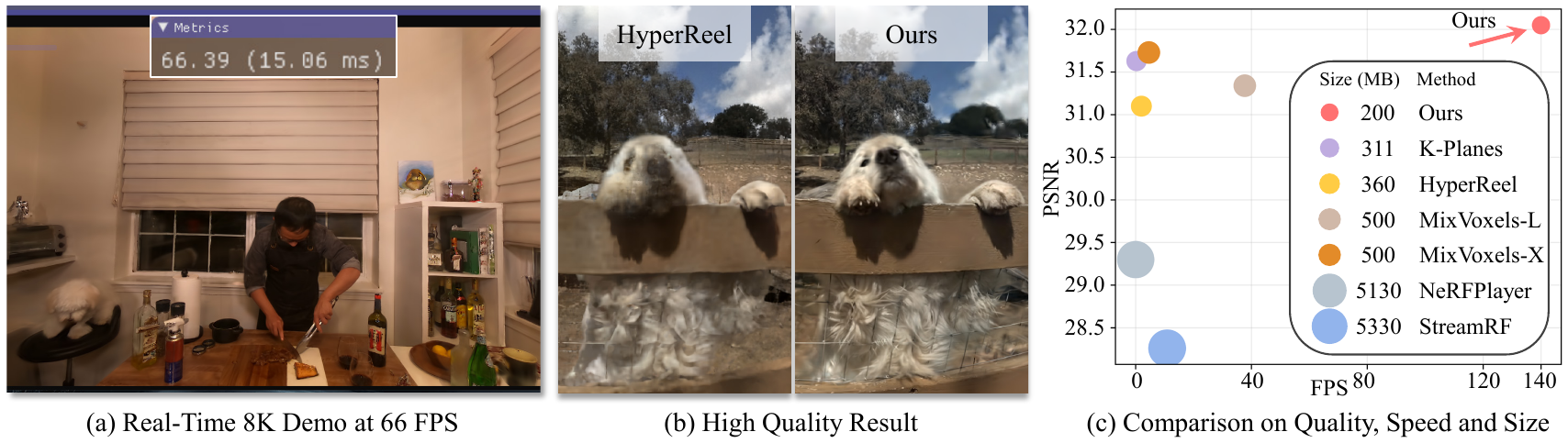}
  \vspace{-5mm}
  \captionof{figure}{\textbf{Our dynamic scene representation achieves photorealistic quality, real-time high-resolution rendering and compact model size.}
  (a) Our lite-version model can render 8K 6-DoF video at 66 FPS on an Nvidia RTX 4090 GPU.
  (b) Example novel view rendering of a challenging scene.
  (c) Quantitative comparisons of rendering quality, speed and model size with prior arts on the Neural 3D Video
Dataset.
  }
  \vspace{-2mm}
  \label{fig:teaser}\vspace{15pt}
}
\makeatother
\maketitle

\begin{abstract}
\vspace{-2mm}
Novel view synthesis of dynamic scenes has been an intriguing yet challenging problem.
Despite recent advancements, simultaneously achieving high-resolution photorealistic results, real-time rendering, and compact storage remains a formidable task.
To address these challenges, we propose Spacetime Gaussian Feature Splatting as a novel dynamic scene representation, composed of three pivotal components.
First, we formulate expressive Spacetime Gaussians by enhancing 3D Gaussians with temporal opacity and parametric motion/rotation.
This enables Spacetime Gaussians to capture static, dynamic, as well as transient content within a scene.
Second, we introduce splatted feature rendering, which replaces spherical harmonics with neural features.
These features facilitate the modeling of view- and time-dependent appearance while maintaining small size.
Third, we leverage the guidance of training error and coarse depth to sample new Gaussians in areas that are challenging to converge with existing pipelines.
Experiments on several established real-world datasets demonstrate that our method achieves state-of-the-art rendering quality and speed, while retaining compact storage.
At 8K resolution, our lite-version model can render at 60 FPS on an Nvidia RTX 4090 GPU.
Our code is available at \href{https://github.com/oppo-us-research/SpacetimeGaussians}{https://github.com/oppo-us-research/SpacetimeGaussians}.
\end{abstract}
\vspace{-2mm}   
{\let\thefootnote\relax\footnote{{
{$^{\dagger}$ Corresponding authors}.
 }}}
 {\let\thefootnote\relax\footnote{{
{$^{*}$} Work done while Zhan was an intern at OPPO US Research Center.
 }}}
 
\section{Introduction}
\label{sec:intro}
Photorealistic modeling of real-world dynamic scenes has been persistently pursued in computer vision and graphics. 
It allows users to freely explore dynamic scenes at novel viewpoints and timestamps, thus providing strong immersive experience, and can vastly benefit applications in VR/AR, broadcasting, education, \etc.

Recent advances in novel view synthesis, especially Neural Radiance Fields (NeRF)~\cite{mildenhall2020nerf}, have greatly improved the convenience and fidelity of static scene modeling from casual multi-view inputs in non-lab environments.
Since then, large quantities of work~\cite{barron2021mip,barron2022mip,sun2022direct,fridovich2022plenoxels,mueller2022instant,chen2022tensorf,barron2023zipnerf,3dg_2023,Chen2023ICCV} have emerged aiming to enhance rendering quality and speed.
Particularly,~\cite{3dg_2023,Chen2023ICCV} propose to use anisotropic radial basis functions as 3D representations, which are highly adaptive to scene structures and boost the modeling of details.
3D Gaussian Splatting (3DGS)~\cite{3dg_2023} further presents an efficient rasterization-based scheme for differentiable volume rendering.
Instead of shooting rays from camera to the scene and sampling points along each ray, 3DGS rasterizes 3D Gaussians onto image plane via splatting, which brings about notable rendering speedup.

Despite the success on static scenes, directly applying the above methods per-frame to dynamic scenes is challenging, due to the overhead in model size and training time.
State-of-the-art dynamic view synthesis methods~\cite{li2022neural,song2023nerfplayer,attal2023hyperreel,fridovich2023k,cao2023hexplane,Wang2023ICCV} adopt a holistic approach where multiple frames are represented in a single model.
NeRFPlayer~\cite{song2023nerfplayer} and HyperReel~\cite{attal2023hyperreel} combine static spatial representations~\cite{mueller2022instant,chen2022tensorf} with temporal feature sharing/interpolation to improve model compactness.
This strategy exploits the characteristic that adjacent frames in natural videos usually exhibit high similarity.
In a similar vein, MixVoxels~\cite{Wang2023ICCV} uses time-variant latents and bridges them with spatial features by inner product.
K-Planes~\cite{fridovich2023k} and HexPlane~\cite{cao2023hexplane} factorize the 4D spacetime domain into multiple 2D planes for compact representation.
One limitation of these methods is that their grid-like representations cannot fully adapt to the dynamics of scene structures, hindering the modeling of delicate details.
Meanwhile, they struggle to produce real-time high-resolution rendering without sacrificing quality.

In this work, we present a novel representation for dynamic view synthesis.
Our approach simultaneously achieves photorealistic quality, real-time high-resolution rendering and compact model size (see \cref{fig:teaser} for example results and comparisons with state-of-the-arts). 
At the core of our approach is Spacetime Gaussian (STG), which extends 3D Gaussian to 4D spacetime domain.
We propose to equip 3D Gaussian with time-dependent opacity along with polynomially parameterized motion and rotation.
As a result, STGs are capable of faithfully modeling static, dynamic as well as transient (\ie, emerging or vanishing) content in a scene.

To enhance model compactness and account for time-varying appearance, we propose splatted feature rendering.
Specifically, for each Spacetime Gaussian, instead of storing spherical harmonic coefficients, we store features that encode base color, view-related information and time-related information.
These features are rasterized to image space via differentiable splatting, and then go through a tiny multi-layer perceptrons (MLP) network to produce the final color.
While smaller in size than spherical harmonics, these features exhibit strong expressiveness.

Additionally, we introduce guided sampling of Gaussians to improve rendering quality of complex scenes. 
We observe that distant areas which are sparsely covered by Gaussians at initialization tend to have blurry rendering results.
To tackle this problem, we propose to sample new Gaussians in the 4D scene with the guidance of training error and coarse depth.

In summary, the contributions of our work are the following:
\begin{itemize}
\item We present a novel representation based on Spacetime Gaussian for high-fidelity and efficient dynamic view synthesis.
\item We propose splatted feature rendering, which enhances model compactness and facilitates the modeling of time-varying appearance.
\item We introduce a guided sampling approach for Gaussians to improve rendering quality at distant sparsely covered areas.
\item Extensive experiments on various real-world datasets demonstrate that our method achieves state-of-the-art rendering quality and speed while keeping small model size. 
Our lite-version model enables 8K rendering at 60 FPS.
\end{itemize}

\section{Related Work}
\label{sec:relatework}

\paragraph{Novel View Synthesis.}

Early approaches leverage image-based rendering techniques with proxy geometry/depth to sample novel views from source images~\cite{Debevec96,gortler1996lumigraph,levoy1996light,Heigl1999,Buehler2001,zheng2009parallax,kopf2014first,ortiz2015bayesian}.
Chaurasia~\etal~\cite{chaurasia2013depth} estimate a depth map to blend pixels from source views and employ superpixels to compensate for missing depth data. 
Hedman~\etal~\cite{Hedman2016} utilize RGBD sensors to improve rendering quality and speed. 
Penner and Zhang~\cite{Penner2017} leverage volumetric voxels for continuity in synthesized views and robustness to depth uncertainty.
Hedman~\etal~\cite{hedman2018deep} learn the blending scheme with neural networks.
Flynn~\etal~\cite{flynn2019deepview} combine multi-plane images with learned gradient descent. Wiles~\etal~\cite{wiles2020synsin} splat latent features from point cloud for novel view synthesis.

\paragraph{Neural Scene Representations.}
In recent years, neural scene representations have achieved great progress in novel view synthesis.
These methods allocate neural features to structures such as volume~\cite{sitzmann2019deepvoxels,Lombardi2019}, texture~\cite{thies2019deferred,chen2020neural}, or point cloud~\cite{aliev2020neural}.
The seminal work of NeRF~\cite{mildenhall2020nerf} proposes to leverage differentiable volume rendering.
It does not require proxy geometry and instead uses MLPs to implicitly encode density and radiance in 3D space.
Later on, numerous works emerge to boost the quality and efficiency of differentiable volume rendering.
One group of methods focuses on improving the sampling strategy to reduce the number of point queries~\cite{neff2021donerf, piala2021terminerf, attal2023hyperreel} or applies light field-based formulation~\cite{li2021neulf,sitzmann2021light,feng2021signet,wang2022r2l,attal2022learning,suhail2022light,li2023relit}. 
Another group trades space for speed by incorporating explicit and localized neural representations~\cite{liu2020neural,chen2021multiresolution,takikawa2021neural,yu2021plenoctrees,reiser2021kilonerf,sun2022direct,fridovich2022plenoxels,xu2022point,chen2022tensorf,mueller2022instant,barron2023zipnerf,hu2023tri}.
Among them, to improve model compactness, Instant-NGP~\cite{mueller2022instant} uses hash grid while TensoRF~\cite{chen2022tensorf} utilizes tensor decomposition.

Recently, 3D Gaussian Splatting (3DGS)~\cite{3dg_2023} proposes to use anisotropic 3D Gaussians as scene representation and presents an efficient differentiable rasterizer to splat these Gaussians to the image plane.
Their method enables fast high-resolution rendering, while preserving great rendering quality.
Similar to 3DGS, NeuRBF~\cite{Chen2023ICCV} leverages anisotropic radial basis functions for neural representation and achieves high-fidelity rendering.
However, the above methods focus on static scene representation.

\paragraph{Dynamic Novel View Synthesis.}
A widely adopted setting for dynamic free-viewpoint rendering is using multi-view videos as input.
Classic methods in this area include~\cite{jain1995multiple,kanade1997virtualized,yang2002real,zitnick2004high,collet2015high,li2017robust,li20184d,li20203d,ding2023full}.
More recently, Broxton~\etal~\cite{broxton2020immersive} use multi-sphere image as bootstrap and then convert it to layered meshes.
Bansal~\etal~\cite{bansal20204d} separate static and dynamic contents and manipulate video with deep network in screen space. 
Bemana~\etal~\cite{bemana2020x} learn a neural network to implicitly map view, time or light coordinates to 2D images.
Attal~\etal~\cite{attal2020} use multi-sphere representations to handle the depth and occlusions in 360-degree videos. 
Lin~\etal~\cite{lin2021deep,lin2023view} propose 3D mask volume to addresses the temporal inconsistency of disocclusions. Neural Volumes~\cite{Lombardi2019} uses an encoder-decoder network to encode images into a 3D volume and decode it with volume rendering.
Lombardi~\etal~\cite{lombardi2021mixture} enhance Neural Volumes by decoding a mixture of dynamic geometric primitives from latent code and skipping samples in empty space for efficient ray marching. 
Extending static NeRF-related representations to dynamic scenes are also being actively explored~\cite{li2022neural,wang2022fourier,li2022streaming,song2023nerfplayer,peng2023representing,wang2023neural,shao2023tensor4d,fridovich2023k,cao2023hexplane,attal2023hyperreel,Wang2023ICCV,icsik2023humanrf,lin2023im4d,wang2024masked,Kim2024Sync}.
DyNeRF~\cite{li2022neural} combines NeRF with time-conditioned latent codes to compactly represent dynamic scenes.
StreamRF~\cite{li2022streaming} accelerates the training of dynamic scenes by modeling the differences of consecutive frames. 
NeRFPlayer~\cite{song2023nerfplayer} decomposes scene into static, new and deforming fields and proposes streaming of feature channels. 
MixVoxels~\cite{Wang2023ICCV} represents scene with a mixture of static and dynamic voxels to accelerate rendering.
HyperReel~\cite{attal2023hyperreel} utilizes sampling prediction network to reduce sampling points and leverages keyframe-based representation.
K-Planes~\cite{fridovich2023k}, HexPlane~\cite{cao2023hexplane} and Tensor4D~\cite{shao2023tensor4d} factorize 4D spacetime domain into 2D feature planes for compact model size.

Another line of research tackles dynamic view synthesis from monocular videos~\cite{li2021neural,pumarola2021d,gao2021dynamic,park2021nerfies,tretschk2021non,wang2021neural,xian2021space,du2021neural,park2021hypernerf,fang2022fast,gao2022monocular,liu2023robust,Li2023CVPR}.
Under this setting, a single camera moves around in the dynamic scene, providing only one observed viewpoint at each timestep.
To address the sparsity of supervision, priors on motion, scene flow or depth are usually introduced.
In this work, we focus on the dynamic representation itself and only consider the setting of multi-view input videos.

Recently, there are several work on this topic that are concurrent to ours~\cite{4k4d2023,4DG3DV,yang2023real,yang2023deformable,wu20234d,xie2023physgaussian,kratimenos2024dynmf,huang2023scgs, lin2023gaussianflow}.
4K4D~\cite{4k4d2023} combines 4D point clouds with K-Planes~\cite{fridovich2023k} and discrete image-based rendering, and uses differentiable depth peeling to train the model.
Luiten~\etal~\cite{4DG3DV} models 4D scene with a set of moving 3D Gaussians, whose positions and rotations are discretely defined at each step.
Their method demonstrates appealing results for 3D tracking, but its rendering quality is less favorable due to flickering artifacts.
Yang~\etal~\cite{yang2023real} leverage 4D Gaussians and 4D spherindrical harmonics for dynamic modelling.
4D Gaussians essentially represent motion with linear model.
Comparatively, our polynomial motion model is more expressive, resulting in higher rendering quality.
Our method also has higher rendering speed than their work.
Yang~\etal~\cite{yang2023deformable} and Wu~\etal\cite{wu20234d} prioritize on monocular dynamic view synthesis and employ deformation fields to deform a set of canonical 3D Gaussians.
For multi-view videos setting, the performance of~\cite{yang2023deformable} is not extensively evaluated while~\cite{wu20234d} depicts inferior rendering quality and speed than ours.

\begin{figure*}[h]
  \includegraphics[width=1.0\textwidth]{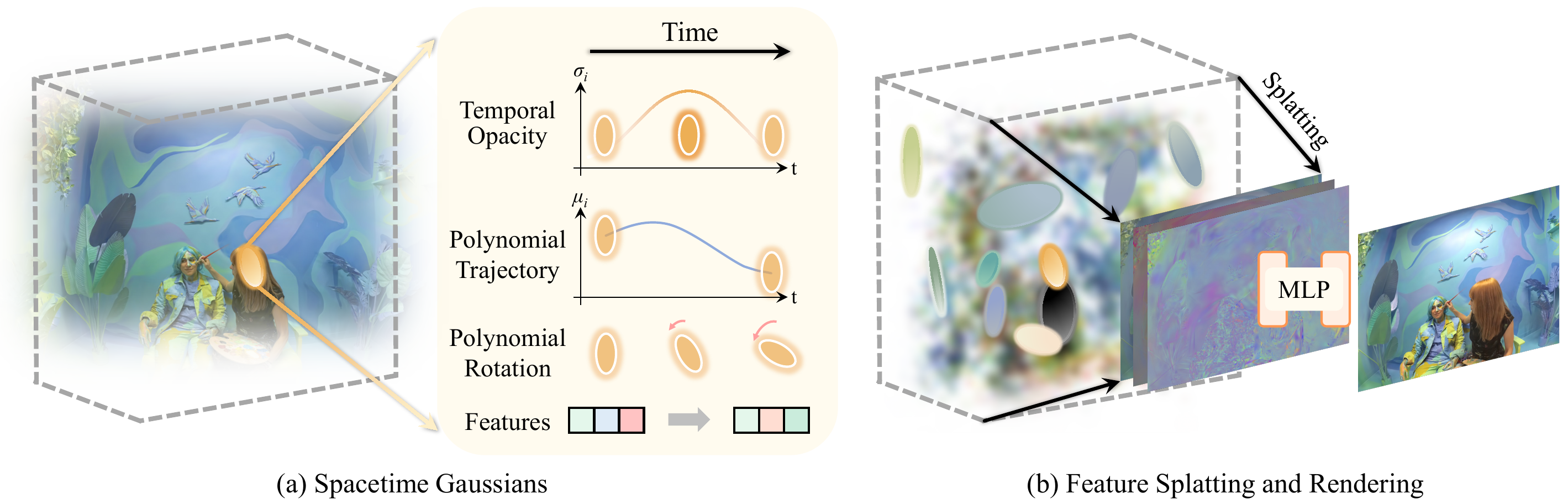}

  \caption{\textbf{Overview of Spacetime Gaussians and splatted feature rendering.} (a) Our method leverages a set of Spacetime Gaussians (STG) to represent the dynamic scenes. On top of 3D Gaussian, each STG is further equipped with temporal opacity, polynomial motion/rotation and time-dependent features. 
  (b) We visualize the splatted features as maps, which are converted to color image via MLP.}
  \label{fig:method}
\end{figure*}

\section{Preliminary: 3D Gaussian Splatting}
Given images at multiple viewpoints with known camera poses, 3D Gaussian Splatting~\cite{3dg_2023} (3DGS) optimizes a set of anisotropic 3D Gaussians via differentiable rasterization to represent a static 3D scene. Owing to their efficient rasterization, the optimized model can render high-fidelity novel views in real-time.

3DGS~\cite{3dg_2023} associates a 3D Gaussian $i$ with a position $\mu_i$, covariance matrix $\Sigma_i$, opacity $\sigma_i$ and spherical harmonics (SH) coefficients $\mathbf{h}_i$. The final opacity of a 3D Gaussian at any spatial point $\mathbf{x}$ is
\begin{equation}
\label{eq:alpha}
\alpha_i = \sigma_i \exp\left( -\frac{1}{2} (\mathbf{x} - {\mu_i})^T \Sigma_{i}^{-1} (\mathbf{x} - \mu_i) \right).
\end{equation}

$\Sigma_{i}$ is positive semi-definite and can be decomposed into scaling matrix $S_i$ and rotation matrix $R_i$:
\begin{equation}\label{eq:covariance}\Sigma_i = R_iS_iS_i^TR_i^T, \end{equation}
where $S_i$ is a diagonal matrix and is parameterized by a 3D vector $\mathbf{s}_i$, and $R_i$ is parameterized by a quaternion $q$.

To render an image, 3D Gaussians are first projected to 2D image space via an approximation of the perspective transformation~\cite{zwicker2001ewa}. Specifically, the projection of a 3D Gaussian is approximated as a 2D Gaussian with center $\mu_i^{2D}$ and covariance $\Sigma_i^{2D}$. Let $W, K$ be the viewing transformation and projection matrix, $\mu_i^{2D}$ and $\Sigma_i^{2D}$ are computed as
\begin{align}
\label{eq:proj_mu}
\mu_i^{2D}&=(K((W\mu_i)/(W\mu_i)_z))_{1:2}, \\
\label{eq:proj_cov}
\Sigma_i^{2D}&=(JW\Sigma_i W^TJ^T)_{1:2,1:2},
\end{align}
where $J$ is the Jacobian of the projective transformation.

After sorting the Gaussians in depth order, the color at a pixel is obtained by volume rendering:
\begin{equation}
	\mathbf{I} = \sum_{i \in \mathcal{N}}
	\mathbf c_{i}\alpha_{i}^{2D}
	\prod_{j=1}^{i-1}(1-\alpha_{j}^{2D}),
 \label{eq:fsplat}
\end{equation}
where $\alpha_{i}^{2D}$ is a 2D version of Eq.~\eqref{eq:alpha}, with $\mu_i, \Sigma_i, \mathbf{x}$ replaced by $\mu_i^{2D}, \Sigma_i^{2D}, \mathbf{x}^{2D}$ (pixel coordinate).
$\mathbf c_{i}$ is the RGB color after evaluating SH with view direction and coefficients $\mathbf{h}_i$.

\section{Method}

We propose a novel representation based on Spacetime Gaussians for modeling dynamic 3D scenes. Our approach takes multi-view videos as input and creates 6-DoF video that allows rendering at novel views.
We first describe the formulation of our Spacetime Gaussian (STG) in~\cref{sec:stg}. 
Then in~\cref{sec:HRender}, we present feature-based splatting for time-varying rendering. \cref{sec:optim} details our optimization process and~\cref{sec:sampling} introduces guided sampling of Gaussians.

\subsection{Spacetime Gaussians}
\label{sec:stg}

To represent 4D dynamics, we propose Spacetime Gaussians (STG) that combine 3D Gaussians with temporal components to model emerging/vanishing content as well as motion/deformation, as shown in~\cref{fig:method} (a).
Specifically, we introduce temporal radial basis function to encode temporal opacity, which can effectively model scene content that emerges or vanishes within the duration of video.
Meanwhile, we utilize time-conditioned parametric functions for the position and rotation of 3D Gaussians to model the motion and deformation in the scene.
For a spacetime point $(\mathbf{x},t)$, the opacity of an STG is
\begin{equation}
\alpha_i(t) = \sigma_i(t) \exp\left( -\frac{1}{2} (\mathbf{x} - {\mu_i(t)})^T \Sigma_i(t)^{-1} (\mathbf{x} - \mu_i(t)) \right),
 \label{eq:stg}
\end{equation}
where $\sigma_i(t)$ is temporal opacity, $\mu_i(t),\Sigma_i(t)$ are time-dependent position and covariance, and $i$ stands for the $i$th STG.
We detail each of the components below.

\paragraph{Temporal Radial Basis Function.}
We use a temporal radial basis function to represent the temporal opacity of an STG at any time $t$. Inspired by \cite{3dg_2023, Chen2023ICCV} that use radial basis functions for approximating spatial signals, we utilize 1D Gaussian for the temporal opacity $\sigma_i(t)$:

\begin{equation}
\label{eq:trbf}
\sigma_i(t) = \sigma_{i}^{s}\exp\left(- s_{i}^{\tau} |t-\mu_{i}^\tau|^{2}\right),
\end{equation}
where $\mu_{i}^\tau$ is temporal center, $s_{i}^\tau$ is temporal scaling factor, and $\sigma_{i}^s$ is time-independent spatial opacity. 
$\mu_{i}^\tau$ represents the timestamp for the STG to be most visible while $s_{i}^\tau$ determines its effective duration (\ie, the time duration where its temporal opacity is high). 
We include $\sigma_{i}^s$ to allow spatial opacity variation across STGs.

\paragraph{Polynomial Motion Trajectory.}
For each STG, we employ a time-conditioned function to model its motion.
Motivated by~\cite{fang2020tpnet,houenou2013vehicle}, we choose polynomial function:
\begin{equation}
\label{eq:motion}
    \mathbf{\mu}_i(t) = \sum_{k=0}^{n_p} b_{i,k} (t - \mu_{i}^\tau)^k,
\end{equation}
where $\mathbf{\mu}_i(t)$ denotes the spatial position of an STG at time $t$. 
$\{ b_{i,k} \}_{k=0}^{n_p}, b_{i,k} \in \mathbb{R}$ are the corresponding polynomial coefficients and are optimized during training.
Combining Eq.~\eqref{eq:trbf} and Eq.~\eqref{eq:motion}, complex and long motion can be represented by multiple short segments with simpler motion. 
In our implementation, we use $n_p=3$ as we find it a good balance between representation capacity and model size. 

\begin{figure*}
  \includegraphics[width=1.0\textwidth]{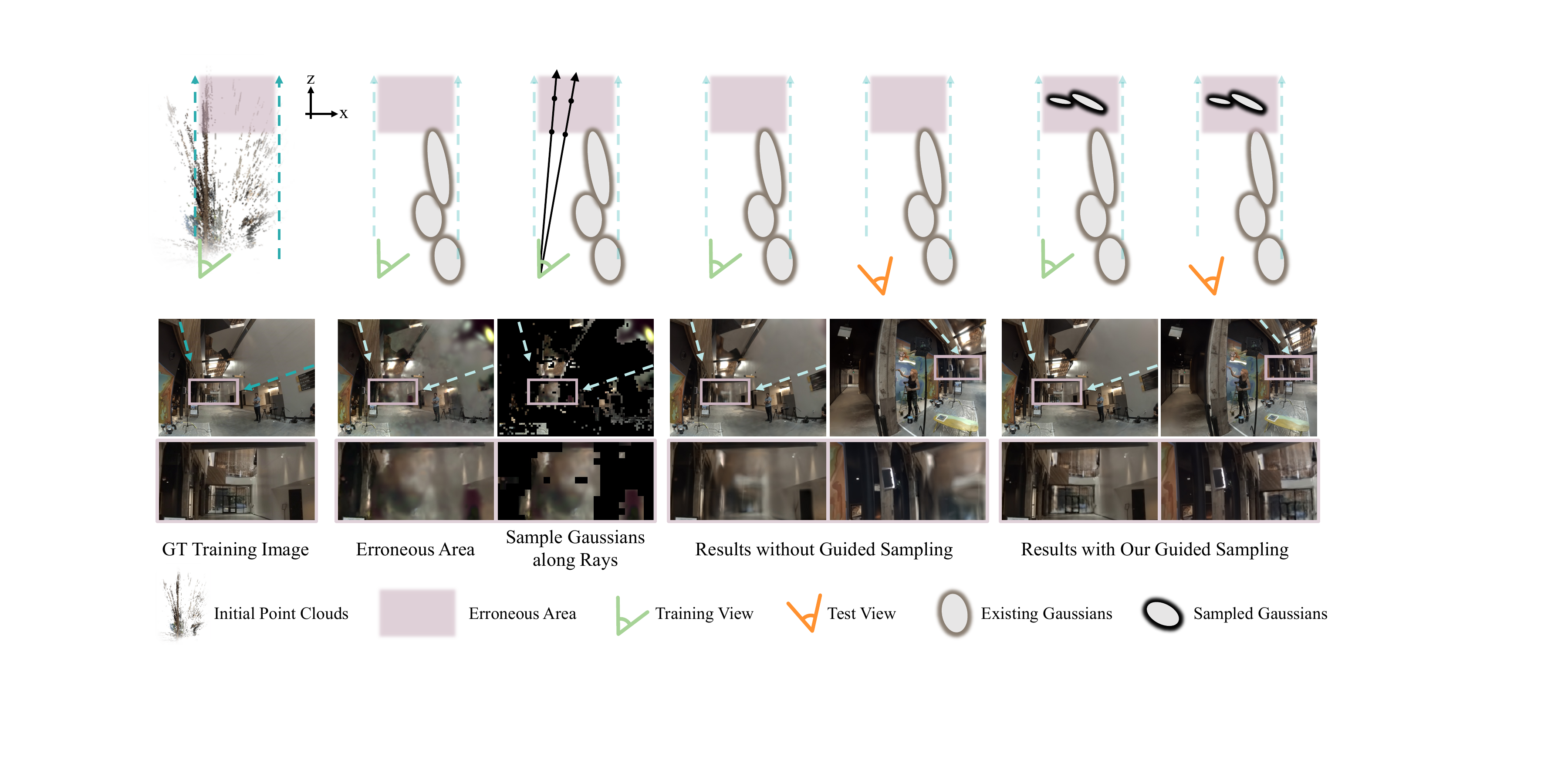}
   \vspace{-0.3in}
  \caption{\textbf{Illustration of our guided sampling strategy for Gaussians.} Our strategy samples new Gaussians along rays by leveraging the guidance of training error and coarse depth.}
  \label{fig:ems}
  \vspace{-5mm}
\end{figure*} 

\paragraph{Polynomial Rotation.}
Following \cite{3dg_2023}, we use real-valued quaternion to parameterize the rotation matrix $R_i$ in Eq.~\eqref{eq:covariance}. 
Similar to motion trajectory, we adopt a polynomial function to represent quaternion:
\begin{equation}
\label{eq:rotation}
    q_i(t) = \sum_{k=0}^{n_q} c_{i,k} (t - \mu_{i}^\tau)^k,
\end{equation}
where $q_i(t)$ is the rotation (in quaternion) of an STG at time $t$, and $\{ c_{i,k} \}_{k=0}^{n_q}, c_{i,k} \in \mathbb{R}$ are polynomial coefficients. 
After converting $q_i(t)$ to rotation matrix $R_i(t)$, the covariance $\Sigma_i(t)$ at time $t$ can be obtained via Eq.~\eqref{eq:covariance}.
We set $n_q=1$ in our experiments.

Note that we keep the scaling matrix $S_i$ in Eq.~\eqref{eq:covariance} to be time-independent, since we experimentally do not observe improvement in rendering quality when applying time-conditioned function on this parameter.

\subsection{Splatted Feature Rendering}
\label{sec:HRender}
To encode view- and time-dependent radiance both accurately and compactly, we store features instead of spherical harmonics coefficients (SH) in each STG. 
Specifically, the features $\mathbf{f}_{i}(t) \in \mathbb{R}^{9}$ of each STG consist of three parts:
\begin{equation}
\label{eq:feature_parts}
    \mathbf{f}_{i}(t) = 
    \begin{bmatrix}
    \mathbf{f}_{i}^{\text{base}}, \mathbf{f}_{i}^{\text{dir}}, & (t-\mu_i^\tau)\mathbf{f}_{i}^{\text{time}}
    \end{bmatrix}^T,
\end{equation}
where $\mathbf{f_{i}^{\text{base}}} \in \mathbb{R}^{3}$ contains base RGB color, and $\mathbf{f_{i}^{\text{dir}}}, \mathbf{f_{i}^{\text{time}}} \in \mathbb{R}^{3}$ encode information related to view direction and time. 
The feature splatting process is similar to Gaussian Splatting~\cite{3dg_2023}, except that the RGB color $\mathbf{c}_i$ in Eq.~\eqref{eq:fsplat} is now replaced by features $\mathbf{f}_{i}(t)$.
After splatting to image space, we split the splatted features at each pixel into $\mathbf{F}^{\textit{base}}, \mathbf{F}^{\textit{dir}}, \mathbf{F}^{\textit{time}}$, whose channels correspond to the three parts in Eq.~\eqref{eq:feature_parts}.
The final RGB color at each pixel is obtained after going through a 2-layer MLP $\Phi$:
\begin{equation}
	\mathbf{I} = \mathbf{F}^{\textit{base}} + \Phi(\mathbf{F}^{\textit{dir}}, \mathbf{F}^{\textit{time}}, \mathbf{r}),
 \label{eq:tdyfsplat}
\end{equation}
where $\mathbf{r}$ is the view direction at the pixel and is additionally concatenated with the features as input. \cref{fig:method} (b) shows an illustration of the rendering process.

Compared to SH encoding, our feature-based approach requires fewer parameters for each STG (9 vs. 48 for 3-degree SH).
At the same time, since the MLP network $\Phi$ is shallow and narrow, our method still achieves fast rendering speed. 

To maximize rendering speed, we can also optionally drop $\Phi$ and only keep $\mathbf{F}^{base}$ during training and rendering. We refer to this configuration as our lite-version model.

\subsection{Optimization}
\label{sec:optim}
The parameters to be optimized include the MLP $\Phi$ and the parameters of each STG ($\sigma_i^s, s_i^\tau, \mu_i^\tau, \{ b_{i,k} \}_{k=0}^{n_p}, \{ c_{i,k} \}_{k=0}^{n_q}, \mathbf{s}_i, \mathbf{f}_i^{\text{base}}, \mathbf{f}_i^{\text{dir}}, \mathbf{f}_i^{\text{time}}$). 

Following~\cite{3dg_2023}, we optimize these parameters through differentiable splatting and gradient-based backpropagation, and interleave with density control of Gaussians.
We use rendering loss that compares rendered images with groundtruth images. The rendering loss consists of a $\mathcal{L}_1$ term and a D-SSIM term.

\subsection{Guided Sampling of Gaussians}
\label{sec:sampling}

We observe that areas which have sparse Gaussians at initialization are challenging to converge to high rendering quality, especially if these areas are far away from the training cameras.
Therefore, we further introduce a strategy to sample new Gaussians with the guidance of training error and coarse depth.

We sample new Gaussians along the rays of pixels that have large errors during training, as illustrated in~\cref{fig:ems}.
To ensure sampling effectiveness, we conduct sampling after training loss is stable.
Since error maps can be noisy during training, we patch-wise aggregate training errors to prioritize on areas with substantial errors rather than outlier pixels.
Then we sample a ray from the center pixel of each selected patches that have large errors.
To avoid sampling in an excessively large depth range, we exploit the coarse depth map of Gaussians' centers to determine a more specific depth range.
The depth map is generated during feature splatting and incurs little computational overhead.
New Gaussians are then uniformly sampled within the depth range along the rays.
We additionally add small noises to the centers of the newly sampled Gaussians.
Among the sampled Gaussians, the unnecessary ones will have low opacity after steps of training and be pruned.
For the scenes in our experiments, the above sampling process only needs to be conducted no more than 3 times.

Our guided sampling strategy is complimentary to the density control techniques in~\cite{3dg_2023}. While density control gradually grows Gaussians near existing ones by splitting, our approach can sample new Gaussians at regions that have sparse or no Gaussians.

\section{Implementation Details}

We initialize our STGs with the structure-from-motion sparse point clouds from all available timestamps. 
For density control, we conduct more aggressive pruning than~\cite{3dg_2023} to reduce the number of Gaussians and keep model size to be relatively small.
We use Adam optimizer~\cite{kingma2014adam}.
The training time for a 50-frame sequence is 40-60 minutes on a single NVIDIA A6000 GPU.
We adapt the splatting process to support different camera models in real world datasets. See supplementary material for more implementation details.

\begin{figure*}[h]
  \includegraphics[width=1.0\textwidth]{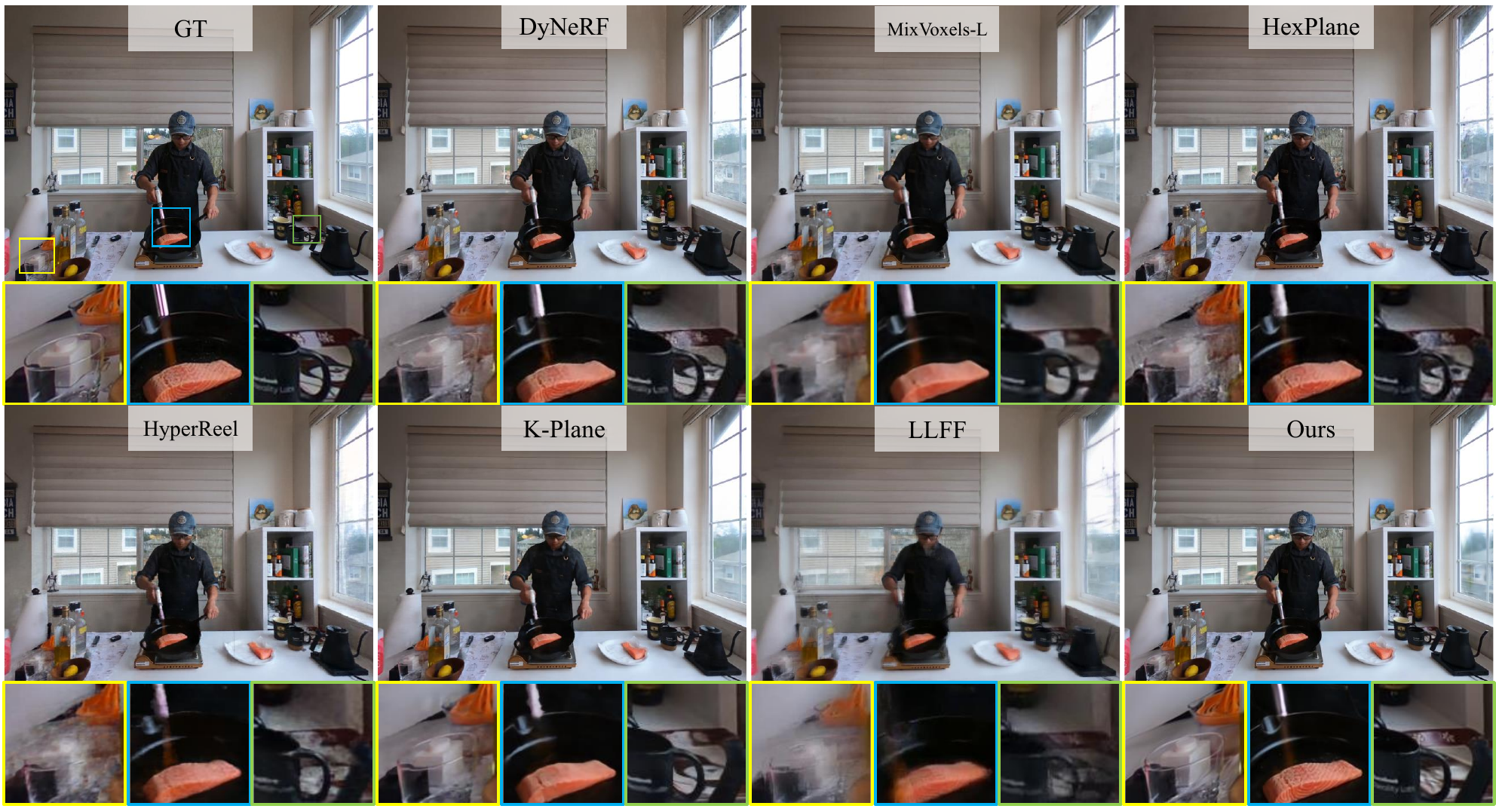}
  % \vspace{-0.3in}
  \caption{\textbf{Qualitative comparisons on the Neural 3D Video Dataset.}}
  \label{fig:comparen4d}
  % \vspace{-0.1in}
\end{figure*}

\section{Experiments}
\label{sec:exp}
We evaluate our method on three real-world benchmarks: Neural 3D Video Dataset~\cite{li2022neural} (\cref{sec:exp_neural3d}), Google Immersive Dataset~\cite{broxton2020immersive} (\cref{sec:exp_immersive}), and Technicolor Dataset~\cite{techni} (\cref{sec:exp_techni}). 
We also conduct ablation studies on various aspects of our method (\cref{sec:exp_ablation}). 
Please refer to supplementary material and video for more results and real-time demo.

\subsection{Neural 3D Video Dataset}
\label{sec:exp_neural3d}

\begin{table}
\caption{\label{tab:n4d}
    \textbf{Quantitative comparisons on the Neural 3D Video Dataset.}
    ``FPS" is measured at \textit{1352 $\times$ 1014} resolution. 
    ``Size" is the total model size for 300 frames. 
    Some methods only report part of the scenes. 
    For fair comparison, we additionally report our results under their settings. 
    $^1$ only includes the \textit{Flame Salmon} scene.
    $^2$ excludes the \textit{Coffee Martini} scene.
}
\vspace{-1mm}
\centering
\resizebox{\linewidth}{!}{
\begin{tabular}{@{}lccccr@{\hspace{0.3\tabcolsep}}r@{}}
    \toprule
   Method & PSNR$\uparrow$ & DSSIM$_1$$\downarrow$& DSSIM$_2$$\downarrow$ & LPIPS$\downarrow$ & FPS$\uparrow$ & Size$\downarrow$ \\
  \midrule
  Neural Volumes~\cite{Lombardi2019}  $^1$ & 22.80 & - & 0.062 & 0.295 & - & - \\
  LLFF~\cite{mildenhall2019llff} $^1$ & 23.24 & - & 0.076 & 0.235 & - & - \\
  DyNeRF~\cite{li2022neural} $^1$   & \textbf{29.58} & - & \textbf{0.020} & 0.083 & 0.015 & \textbf{28 MB} \\ 
  Ours $^1$ & 29.48 & 0.038& 0.022 & \textbf{0.063} & \textbf{103}  & 300 MB \\
    
  \hline
  HexPlane~\cite{cao2023hexplane} $^2$  & 31.71 & - & -  & 0.075 & - & 200 MB \\
  Ours $^2$ & \textbf{32.74} & 0.027 & 0.012  & \textbf{0.039} & \textbf{140} & \textbf{190 MB} \\
  \hline
  StreamRF~\cite{li2022streaming}  & 28.26 & - & - & - & 10.9 & 5310 MB \\
  NeRFPlayer~\cite{song2023nerfplayer}  & 30.69 & 0.034 & - & 0.111 & 0.05 & 5130 MB \\
  HyperReel~\cite{attal2023hyperreel}  & 31.10 & 0.036 & - & 0.096 & 2 & 360 MB \\
  K-Planes~\cite{fridovich2023k}    &  31.63 & - &0.018  & - & 0.3 & 311 MB \\
  MixVoxels-L~\cite{Wang2023ICCV}   & 31.34 &  -  & 0.017   & 0.096 & 37.7  & 500 MB \\
  MixVoxels-X~\cite{Wang2023ICCV}   & 31.73 &  - &0.015   & 0.064 & 4.6 & 500 MB \\
  Ours         & \textbf{32.05} & \textbf{0.026}  & \textbf{0.014} & \textbf{0.044}  & \textbf{140} & \textbf{200 MB} \\
  \bottomrule
\end{tabular}
}
\end{table}

The Neural 3D Video Dataset~\cite{li2022neural} contains six indoor multi-view video sequences captured by 18 to 21 cameras at $2704\times2028$ resolution. 
Following common practice, training and evaluation are conducted at half resolution, and the first camera is held out for evaluation~\cite{li2022neural}.
The number of frames is 300 for each scene. 

We use PSNR, DSSIM and LPIPS~\cite{zhang2018perceptual} as evaluation metrics.
As mentioned in ~\cite{attal2023hyperreel, fridovich2023k}, there is an inconsistency in the DSSIM implementation across methods. 
For fair comparison, we do our best to group existing methods' DSSIM results into two categories (DSSIM$_1$ and DSSIM$_2$).
Using the $structural\_similarity$ function from $scikit$-$image$ library, DSSIM$_1$ sets $data\_range$ to $1.0$ while DSSIM$_2$ sets $data\_range$ to $2.0$.
We use FPS as metric for rendering speed.
Metrics are averaged over all six scenes except noted otherwise. 

As shown in~\cref{tab:n4d}, our method achieves 140 FPS and outperforms the others by a large margin. 
Our approach also has the best LPIPS in all comparisons and the best PSNR/DSSIM in most cases.
\cref{fig:comparen4d} shows qualitative comparisons on a representative view that is widely used in other work.
Compared to the other baselines, our result contains more vivid details (\eg, the textures on the salmon) and artifact-less rendering (\eg, the caustics on the cup).
Please see supplementary material for comparisons with concurrent methods.

\subsection{Google Immersive Dataset}
\label{sec:exp_immersive}

\begin{table}
\caption{\label{tab:immersive}%
    \textbf{Quantitative comparisons on the Google Immersive Dataset.}
    ``Size/Fr" stands for model size per frame.
}
 \vspace{-1mm}
\renewcommand{\arraystretch}{1}
\centering
\resizebox{\linewidth}{!}{%
\begin{tabular}{@{}lcccr@{\hspace{0.75\tabcolsep}}r@{}}
    \toprule
   Method & PSNR$\uparrow$ & DSSIM$_1$$\downarrow$ & LPIPS$\downarrow$ & FPS$\uparrow$ & Size/Fr$\downarrow$ \\
  \midrule
  NeRFPlayer~\cite{song2023nerfplayer} & 25.8 &  0.076  & 0.196 & 0.12 & 17.1 MB \\
  HyperReel~\cite{attal2023hyperreel} &  28.8 &  0.063  & 0.193 & 4 &  \bf 1.2 MB \\
  Ours& \bf 29.2 & \bf 0.042 & \bf 0.081 & \bf 99 & \bf 1.2 MB \\
  \bottomrule
\end{tabular}%
}
\end{table}

\begin{figure*}[h]
  \includegraphics[width=1.0\textwidth]{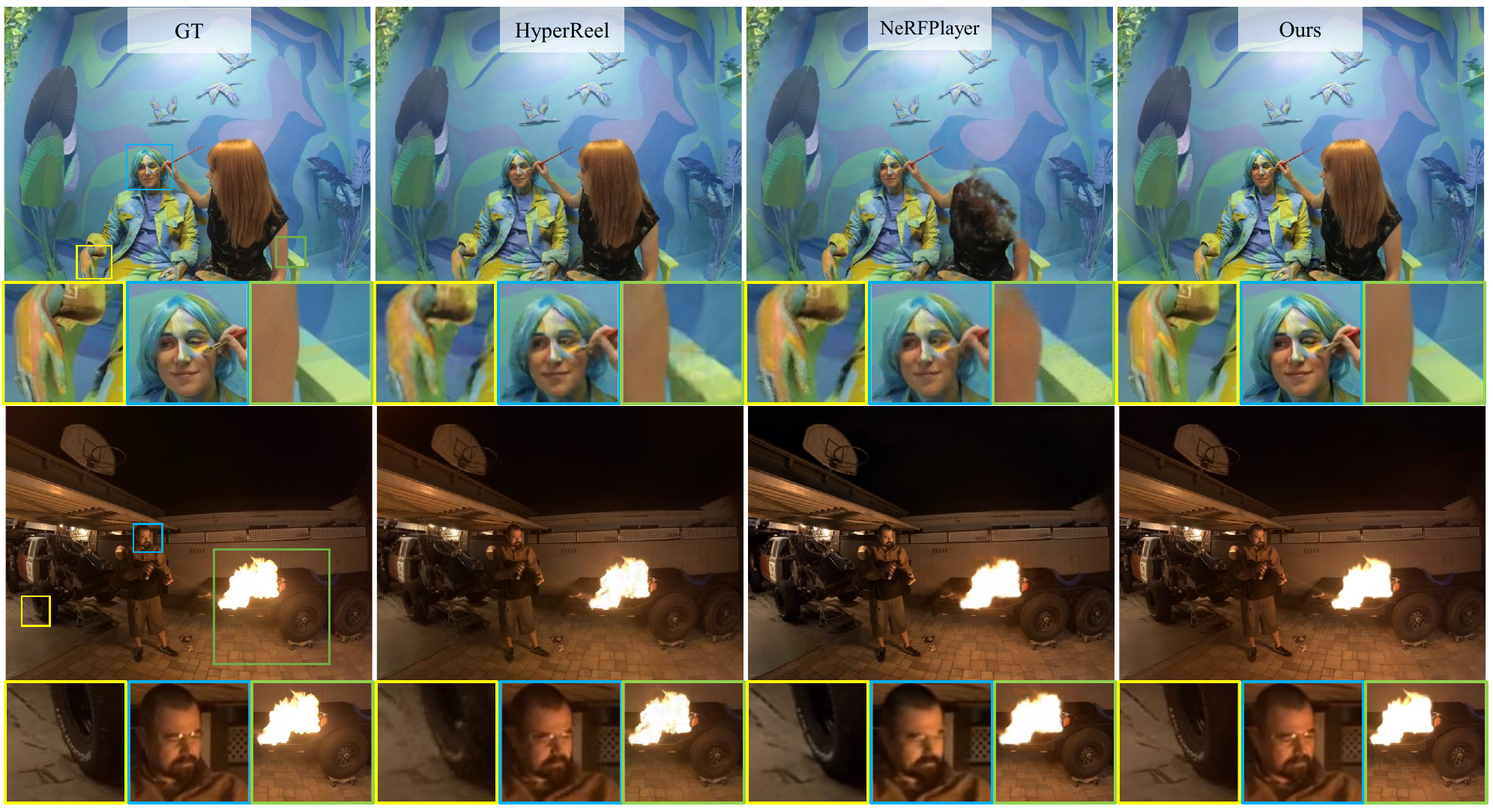}

  \caption{\textbf{Qualitative comparisons on the Google Immersive Dataset.}}
  \label{fig:exampleimmersive}
  \vspace{-3mm}
\end{figure*} 

Google Immersive Dataset~\cite{broxton2020immersive} contains indoor and outdoor scenes captured with a 46-camera rig. 
The cameras are in fish-eye mode and are mounted on an outward-facing hemisphere.
Compared to outside-in setups, there is less overlap among views, hence posing additional challenges.

Following~\cite{song2023nerfplayer,attal2023hyperreel}, we evaluate on 7 selected scenes (\textit{Welder}, \textit{Flames}, \textit{Truck}, \textit{Exhibit}, \textit{Face Paint 1}, \textit{Face Paint 2}, \textit{Cave}) and hold out the center camera as test view.
The numerical results of NeRFPlayer~\cite{song2023nerfplayer} and HyperReel~\cite{attal2023hyperreel} are from their papers. 
The visual results of NeRFPlayer~\cite{song2023nerfplayer} are obtained from their authors while those of HyperReel~\cite{attal2023hyperreel} are produced by running their released codes.
 
As shown in~\cref{tab:immersive}, our method outperforms NeRFPlayer and HyperReel in both speed and quality. 
Compared to HyperReel, our method is over 10 times faster in rendering speed.
Although our PSNR is only 0.4 dB higher, the improvements in DSSIM and LPIPS are significant.
When compared to NeRFPlayer, the margin is larger for all metrics.
Visual comparisons are shown in~\cref{fig:exampleimmersive}.
Our method depicts sharper details and fewer artifacts than the others.

\subsection{Technicolor Dataset}
\label{sec:exp_techni}

\begin{table}
\caption{\label{tab:quant_dynamic}%
    \textbf{Quantitative comparisons on the Technicolor Dataset.}
    ``Size/Fr" stands for model size per frame.
}
 \vspace{-1mm}
\renewcommand{\arraystretch}{1}
\centering
\resizebox{\linewidth}{!}{%
\begin{tabular}{@{}lccccr@{\hspace{0.75\tabcolsep}}r@{}}
    \toprule
   Method & PSNR$\uparrow$ & DSSIM$_1$$\downarrow$ & DSSIM$_2$$\downarrow$ & LPIPS$\downarrow$ & FPS$\uparrow$ & Size/Fr$\downarrow$ \\
  \midrule
  DyNeRF~\cite{li2022neural} & 31.8 & - & 0.021 & 0.140 & 0.02 & \bf 0.6 MB \\
  HyperReel~\cite{attal2023hyperreel} &  32.7 & 0.047 & - & 0.109 & 4.00 & 1.2 MB \\
  Ours  & \bf 33.6 & \bf 0.040 & \bf 0.019 & \bf 0.084 & \bf 86.7 & 1.1 MB \\
  \bottomrule
\end{tabular}%
\label{tab:technicolor}
}
\vspace{-2mm}
\end{table}

Technicolor Light Field Dataset~\cite{techni} contains videos taken with a 4x4 camera array. 
Each camera is time-synchronized and the spatial resolution is $2048\times1088$. 
In alignment with HyperReel~\cite{attal2023hyperreel}, we hold out the camera at second row second column and evaluate on five scenes (\textit{Birthday}, \textit{Fabien}, \textit{Painter}, \textit{Theater}, \textit{Trains}) at full resolution.
 
~\cref{tab:technicolor} shows the comparisons, where our method achieves noticeable gain in quality and speed.
Please refer to supplementary material for visual comparisons.
 
\subsection{Ablation Study}
\label{sec:exp_ablation}

\begin{table}
\caption{\label{tab:techniab}%
    \textbf{Ablation study of proposed components.} Conducted on all the five scenes from the Technicolor Dataset.
}
\vspace{-2mm}
\renewcommand{\arraystretch}{1}
\centering
\resizebox{\linewidth}{!}{%
\begin{tabular}{@{}lcccr@{\hspace{0.75\tabcolsep}}r@{}}
    \toprule
   Method & PSNR$\uparrow$ & DSSIM$_1$$\downarrow$ & LPIPS$\downarrow$ \\
  \midrule

  w/o Temporal Opacity & 31.0  & 0.063  & 0.153 \\
  w/o Polynomial Motion          & 32.6  & 0.045  & 0.099 \\
  w/o Polynomial Rotation            & 33.4  & 0.042  & 0.085 \\  
  w/o Feature Splatting & 33.0  & 0.044  & 0.097 \\
  w/o Guided Sampling of Gaussians          & 33.3  & 0.041  & 0.085 \\
  
  Ours Full                         & \bf 33.6   & \bf 0.040 & \bf 0.084  \\
  \bottomrule
\end{tabular}%
}  
\vspace{-2mm}
\end{table}

\begin{table}
\caption{\label{tab:techniabsize}%
    \textbf{Ablation study on the number of frames whose SfM point clouds are used in initialization.}
    Conducted on the \textit{Theater} scene from the Technicolor Dataset.
}
% \vspace{-1mm}
\renewcommand{\arraystretch}{1}
\centering
\resizebox{\linewidth}{!}{%
\begin{tabular}{@{}lcccr@{\hspace{0.75\tabcolsep}}r@{}}
    \toprule
   Every N Frames & PSNR$\uparrow$ & DSSIM$_1$$\downarrow$ & LPIPS$\downarrow$ & Size$\downarrow$ \\
  \midrule

  N = 1      & \bf 31.58  &0.059  &0.124  & 110.2 MB \\
  N = 4      & 31.51  & \bf 0.057  & \bf 0.117  & 46.7 MB  \\  
  N = 16     & 31.04  &0.060  &0.139  & \bf 32.6 MB  \\  
\bottomrule
\end{tabular}%
}  
\vspace{-3mm}
\end{table}

To evaluate the effectiveness of proposed components, we conduct an ablation study in~\cref{tab:techniab} using all the five scenes from Technicolor Dataset. Below we describe the configuration and performance of each ablation baseline.

\noindent\textbf{Temporal Opacity.} ``w/o Temporal Opacity" fixes the center and scale of the temporal radial basis functions during training. This variant suffers from a significant performance drop, revealing the importance of temporal opacity.

\noindent\textbf{Polynomial Motion and Rotation.}
``w/o Polynomial Motion" and ``w/o Polynomial Rotation" fix the spatial position and rotation of STGs respectively.
Both lead to a performance drop.
Comparatively, motion is more important than rotation, which motivates us to use a lower-degree polynomial for rotation.

\noindent\textbf{Feature Splatting.}
``w/o Feature Splatting" uses the base RGB color $\mathbf{F}^{base}$ as the final color.
It can be seen that there is a moderate drop in quality due to reduced ability to model view- and time-dependent appearance.

\noindent\textbf{Guided Sampling of Gaussians.}
``w/o Guided Sampling of Gaussians" does not encounter much performance drop in this dataset.
The reason is that the scenes contain rich textures and can be well covered by SfM points.
However, for other challenging scenes, guided sampling plays an important role (see~\cref{fig:ems} and supplementary material for examples).

\noindent\textbf{Number of Frames used for Initialization.}
We further analyzed the number of frames used for initialization in~\cref{tab:techniabsize}.
Using fewer frames slightly downgrade quality, but also significantly reduces model size.
It reveals that the compactness of our method can be further enhanced with a good selection of frames for initialization.

\subsection{Limitations}
Although our representation achieves fast rendering speed, it cannot be trained on-the-fly.
The support for on-the-fly training could benefit numerous streaming applications.
To achieve this, advanced initialization techniques could be explored to accelerate the training process or alleviate the requirement of per-scene training.
On the other hand, our method currently focuses on multi-view video inputs. 
It is promising to adapt our approach to monocular setting by combining with regularization or generative priors.

\section{Conclusion}
We present a novel representation based on Spacetime Gaussians for dynamic view synthesis. 
The proposed Spacetime Gaussians are enhanced with temporal opacity and parametric motion/rotation to model complex 4D content. 
To increase model compactness and encode view/time-dependent appearance, we introduce splatted feature rendering, which utilizes neural features and a lightweight MLP instead of spherical harmonics.
Additionally, we leverage guided sampling of Gaussians to further improve the rendering quality of complex scenes.
Experiments on real-world datasets show that our representation delivers state-of-the-art quality at high resolution and FPS, while maintaining a compact model size.

{
    \small
    \bibliographystyle{ieeenat_fullname}
    \bibliography{citation}
}

\clearpage

\appendix

\section{Overview}
\label{sec:Supplementary material oveview}
Within the supplementary material, we provide:
\begin{itemize}
\item Quantitative and qualitative comparisons to concurrent work in~\cref{sec::concurrentwork}.

\item More ablation study in \cref{sec::moreabalations}.

\item Additional discussions in \cref{sec::additionaldisscusion}.

\item Additional experiment details in \cref{sec::expdetails}.

\item Per-scene quantitative comparisons and more visual comparisons with other methods on the Neural 3D Video Dataset~\cite{li2022neural}, Google Immersive Dataset~\cite{broxton2020immersive} and Technicolor Dataset~\cite{techni} in~\cref{sec::supplevisual}.

\item Real-time demos and dynamic comparisons in our video. Please refer to our website. 

\end{itemize}

\section{Comparisons with Concurrent Work}
\label{sec::concurrentwork}
We compare with concurrent work~\cite{4DG3DV,yang2023real,wu20234d,4k4d2023} on the Neural 3D Video Dataset~\cite{li2022neural} in~\cref{tab:n4dall}. 
We also include Im4D~\cite{lin2023im4d} in this comparison since it is related to 4K4D~\cite{4k4d2023}.
Same with~\cref{tab:n4d} in the main paper, we group DSSIM results into two categories (DSSIM$_1$: $data\_range$ is set to $1.0$; DSSIM$_2$: $data\_range$ is set to $2.0$).

Compared to methods~\cite{4DG3DV,yang2023real,wu20234d} that similarly build upon Gaussian Splatting, our method achieves the best rendering quality and is among the fastest and most compact ones.
Specifically, in terms of quality, our full model performs the best on all of PSNR, DSSIM and LPIPS.
Meanwhile, our lite model also outperforms Dynamic 3DGS~\cite{4DG3DV} and 4DGaussians~\cite{wu20234d} by a noticeable margin, and is only inferior to 4DGS~\cite{yang2023real}.

\begin{figure*}
  \includegraphics[width=1.0\textwidth]{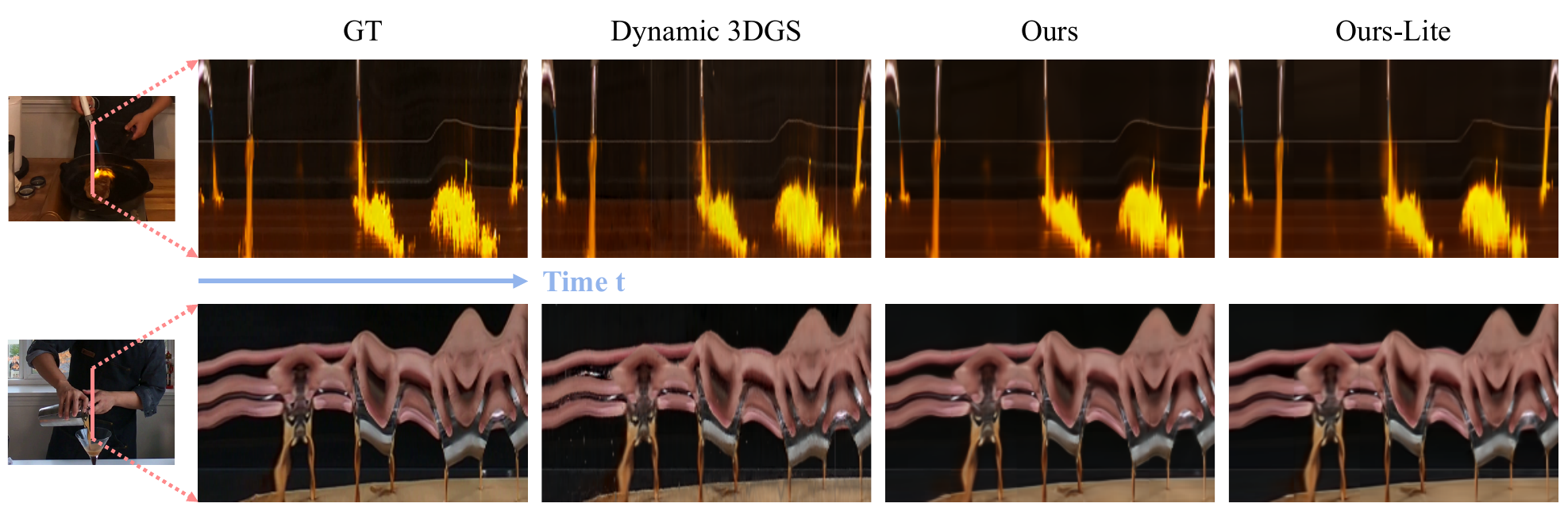}
  % \vspace{-0.3in}
  \caption{\textbf{Comparisons of temporal consistency on the Neural 3D Video Dataset}. 
  From the test view video results of each method, we take a vertical column of 150 pixels across 250 frames and concatenate these columns horizontally.
  The resulting image patch is equivalent to a slice in the height-time space.
  Ours results are clearer than Dynamic 3DGS~\cite{4DG3DV} and contain fewer temporal noises.
  }
  \label{fig:comp_temporal}
  % \vspace{-0.1in}
\end{figure*}

Both our lite model and Dynamic 3DGS~\cite{4DG3DV} can run at over $300$ FPS on the Neural 3D Video Dataset. 
Although our full model is slower than these two, it is still faster than 4DGS~\cite{yang2023real} and 4DGaussians~\cite{wu20234d}.
Compared with Dynamic 3DGS, our lite model takes about only six percent of model size and is $0.6$ dB higher in PSNR.
Meanwhile, the results of Dynamic 3DGS contain many time-varying floaters, which harm temporal consistency and visual quality.
To illustrate this, we show the slices of a column of pixels across time in~\cref{fig:comp_temporal}.
In this visualization, temporal noises appear as sharp vertical lines or dots.
It can be seen that the results of Dynamic 3DGS contain many such patterns.
On the contrary, our results are free of these artifacts.
One reason for this phenomenon is that Dynamic 3DGS requires per-frame training, while ours trains across a sequence of frames.
As a result, our method can better preserve the temporal consistency across frames.
Please refer to our video for dynamic comparisons. 

Compared to Im4D~\cite{lin2023im4d} and 4K4D~\cite{4k4d2023}, both our full model and lite-version model achieve higher rendering quality and speed.

\begin{table*}
\caption{\label{tab:n4dall}
    \textbf{Quantitative comparisons on the Neural 3D Video Dataset.}
    ``FPS" is measured at \textit{1352 $\times$ 1014} resolution. 
    ``Size" is the total model size for 300 frames. 
    Some methods only report results on part of the scenes. 
    For fair comparison, we additionally report our results under their settings. 
    $^3$ only includes the \textit{Cook Spinach}, \textit{Cut Roasted Beef}, and \textit{Sear Steak} scenes.
    $^4$ only includes the \textit{Cut Roasted Beef} scene.
    For the LPIPS metric, no annotation means LPIPS$_{Alex}$, $^{V}$ denotes LPIPS$_{VGG}$. $^\dagger$ denotes it is unclear which LPIPS or DSSIM is used from the corresponding paper.
 }
% \vspace{-1mm}
% \renewcommand{\arraystretch}{1}
\centering
\resizebox{0.7\linewidth}{!}{%
\begin{tabular}{@{}lccccr@{\hspace{0.3\tabcolsep}}r@{}}
\toprule
   Method & PSNR$\uparrow$ & DSSIM$_1$$\downarrow$& DSSIM$_2$$\downarrow$ & LPIPS$\downarrow$ & FPS$\uparrow$ & Size$\downarrow$ \\
\midrule
  Dynamic 3DGS~\cite{4DG3DV}  & 30.67 & 0.035 & 0.019 & 0.099 & \textbf{460} & 2772 MB \\
4DGaussians~\cite{wu20234d}  &31.15 & - & 0.016 $^\dagger$ & 0.049 $^\dagger$ & 30 & \bf 90MB \\
   
  4DGS~\cite{yang2023real}  & 32.01 & - & \bf 0.014 & 0.055 & 114 & - \\
  Ours         & \textbf{32.05} & \textbf{0.026}  & \textbf{0.014} & \textbf{0.044} & 140 & 200 MB \\
   Ours-Lite    & 31.59 & 0.027  &0.015 & 0.047 & 310 &  103 MB \\
\midrule
   4DGaussians~\cite{wu20234d} $^3$ & 32.62  & 0.023 $^\dagger$ & - & -  & - & - \\
    
   Ours $^3$ & \bf 33.53 & \bf 0.020 & \bf 0.010 & \bf 0.034, \bf 0.131 $^{V}$ & 154 & 148MB \\
   Ours-Lite $^3$  & 33.36 & \bf 0.020 & 0.011 & 0.036, 0.133 $^{V}$ & \textbf{330} & \bf 83MB \\
\midrule
  Im4D~\cite{lin2023im4d} $^4$  & 32.58 & - & 0.015 & 0.208 $^{V}$ & - & - \\ 
  4K4D~\cite{4k4d2023} $^4$  & 32.86 & - & 0.014  & 0.167 $^{V}$ & 110 & - \\
  Ours $^4$   & 33.52 & \bf 0.020 & \bf 0.011 & \bf 0.035, \bf 0.133 $^{V}$ & 151 & 154 MB \\ 
  Ours-Lite $^4$   & \bf 33.72 & 0.021 & \bf 0.011 & 0.038, 0.136 $^{V}$& \textbf{338} & \textbf{80 MB} \\ 
\bottomrule
\end{tabular}
}
\end{table*}

\begin{figure*}
  \includegraphics[width=1.0\textwidth]{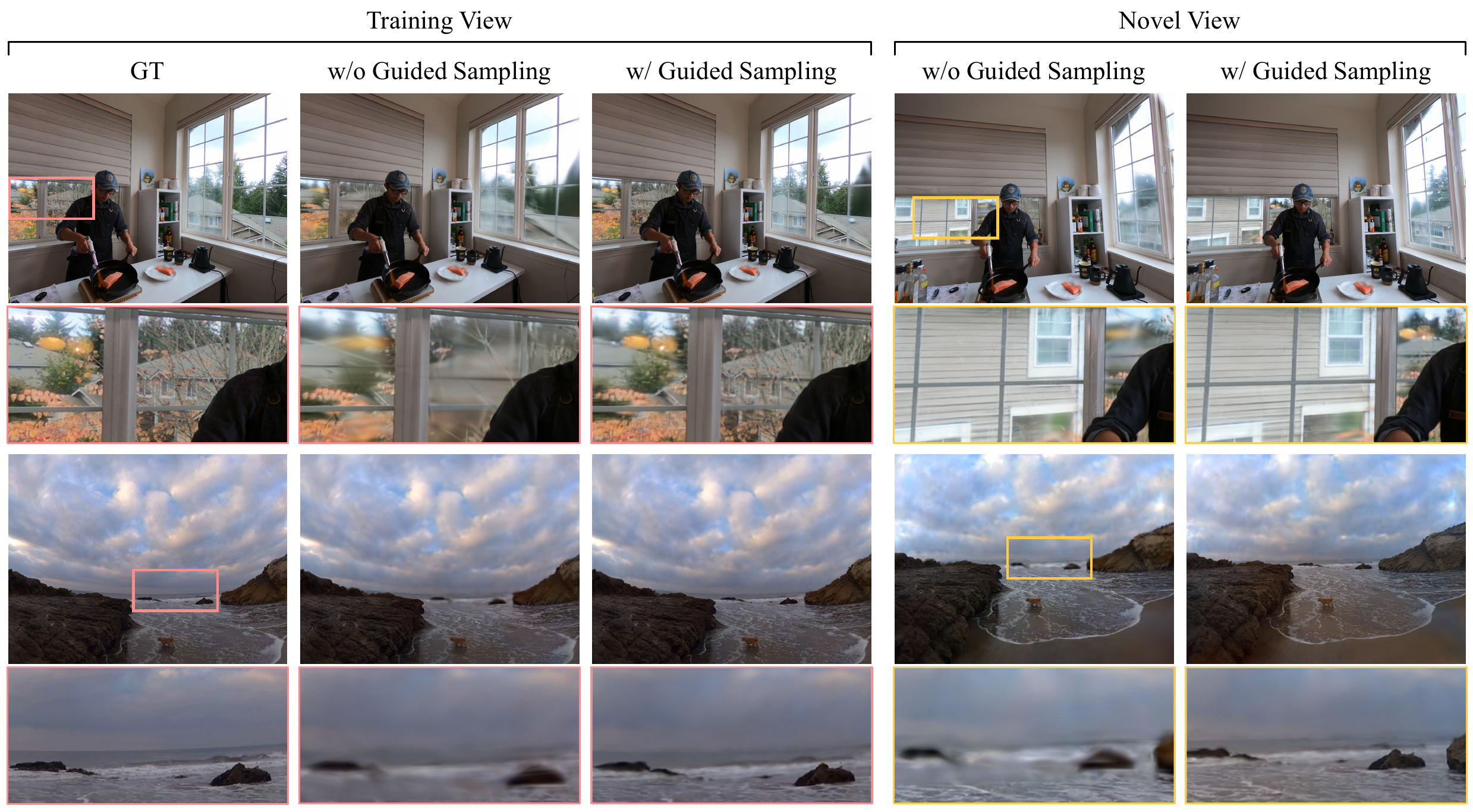}
  % \vspace{-0.3in}
  \caption{\textbf{Ablation on Guided Sampling.} With guided sampling, the rendering results contain less blurriness in both training and novel views.}
  \label{fig:comp_guided_samp}
  % \vspace{-0.1in}
\end{figure*}

\section{More Ablation Study}
\label{sec::moreabalations}
\subsection{Guided Sampling and Strategies of Adding Gaussians}

We visualize the effects of guided sampling in~\cref{fig:comp_guided_samp}. 
It can be seen that when without guided sampling, distant areas that are not well covered by SfM points will have very blurry rendering in both training and novel views.
It reveals that it is challenging to pull Gaussians to these areas with gradient-based optimization and density control.
On the other hand, with guided sampling applied, the renderings at these areas become much sharper for both training and novel views.
Note that the color tone difference in the bottom two rows is caused by inconsistent white balance in the training views of the scene, which makes each model have slightly different color tone in the novel view.

We also compare our guided sampling with two other strategies. The first one randomly adds Gaussians in the whole space and the second one adds a sphere of Gaussians near the far points of our guided sampling.
As shown in~\cref{tab:guidedsamling} rows 2-5, our method has over $0.7$dB PSNR improvement.

\subsection{Analysis on More Scenes}

~\cref{tab:guidedsamling} rows 4-9 extend the ablation study in~\cref{tab:techniab} of the main paper to additional scenes from the Neural 3D Video Dataset and the Google Immersive Dataset. We can see that our proposed components remain effective under various camera setup and scene content.

\subsection{Polynomial Orders and Replacing Polynomials with MLP}

In this experiment, we alter the polynomial orders $n_p, n_q$ and replace the polynomials with MLPs. ~\cref{tab:changeNpNq} shows the results. Our choice of $n_p, n_q$ and polynomials balances quality and storage. 

\begin{table}
\caption{\textbf{Ablation of guided sampling and other components.} Conducted on the first 50 frames of \textit{Flame Salmon} and \textit{09\_Exhibit} scenes.}
\label{tab:guidedsamling}
% \vspace{-0.5mm}
\begin{center}
\renewcommand{\arraystretch}{0.65}
\resizebox{1.0\columnwidth}{!}{
\begin{tabular}{l|cccc}
\toprule
 & PSNR$\uparrow$ & DSSIM$_1$$\downarrow$ & LPIPS$\downarrow$ \\
\midrule

Add random Gaussians during init  &27.72	&0.0455	&0.0787

  \\

Add a sphere of Gaussians during init     &29.13	&0.0381	&0.0690 \\

w/o Guided Sampling               &27.48	&0.0453	&0.0921 \\
\midrule

Ours-Full                 &\textbf{29.88}	&\textbf{0.0373}	&\textbf{0.0665} \\

\midrule
w/o Temporal Opacity  &28.82	&0.0376	&0.0673
 \\
w/o Polynomial Motion     &28.35	&0.0406	&0.0688
  \\
w/o Polynomial Rotation   &28.69	&0.0455	&0.0690
 \\
w/o Feature Splatting &28.05	&0.0448	&0.0754
 \\

\bottomrule
\end{tabular}
}
\end{center}
\end{table}

\begin{table}[h]
\caption{\textbf{Ablation of temporal function, polynomial orders, and features.} Conducted on the first 50 frames of \textit{Theater} and \textit{Sear Steak} scenes.}
\label{tab:changeNpNq}
% \vspace{-2.5mm}
\begin{center}
\renewcommand{\arraystretch}{0.2}
\resizebox{1.0\columnwidth}{!}{
\begin{tabular}{lcccc}
\toprule
   & Size (MB)$\downarrow$  & PSNR$\uparrow$ & DSSIM$_1$$\downarrow$ & LPIPS$\downarrow$ \\
\midrule
Ours-Temporal-MLP   &41.6	&30.93	&0.0428	&0.0967
\\

\midrule
$n_{p}=1$  &33.4	&32.24	&0.0388	&0.0823

 \\

$n_{p}=2$  &37.3	&32.43	&0.0379	&0.0820

 \\
$n_{p}=4$  &44.3	&\textbf{32.61}	&\textbf{0.0374}	&\textbf{0.0809}

 \\
$n_{q}=2$  &45.0	&32.60	&0.0377	&0.0813

 \\

\midrule

Ours-Full ($n_{p}=3,n_{q}=1$)    &40.7	&32.56	&0.0376	&0.0816 \\
\midrule

w/o  $\mathbf{f}^{base}$   &\textbf{31.9} &31.83	&0.0408	&0.0959
\\
w/o $\mathbf{f}^{dir}$    &38.5	&32.03	&0.0384	&0.0849 \\
w/o $\mathbf{f}^{time}$  &39.1	&32.03	&0.0392	&0.0818
 \\
\midrule

Random init $\mathbf{f}^{base}$ &36.3	&32.14	&0.0385	&0.0841
  \\
Random init $\mathbf{f}^{dir}$  &40.7	&32.22	&0.0379	&0.0827
 \\
Random init $\mathbf{f}^{time}$  &40.8	&32.45	&0.0378	&0.0814
 \\

\bottomrule
\end{tabular}
}
\end{center}
\end{table}

\subsection{Feature Components}
\begin{figure*}
  \includegraphics[width=1\linewidth]{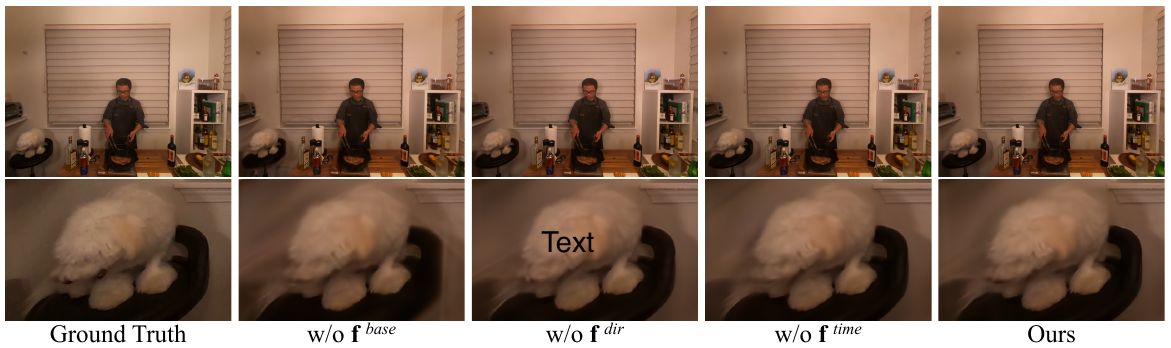}
  % \vspace{-0.3in}
  \caption{\textbf{Ablation on feature components.} 
  Using all features produces the best visual quality.
  }
  \label{fig:comp_features}
  % \vspace{-0.1in}
\end{figure*}

In this experiment, we ablate different features ($\mathbf{f}^{base}$, $\mathbf{f}^{dir}$ and $\mathbf{f}^{time}$) used in our full model. 
As shown in~\cref{tab:changeNpNq} rows 7-10 and~\cref{fig:comp_features}, each component boosts rendering quality.

\begin{figure*}
 % \vspace{-1mm}
  \centering
  \includegraphics[width=1.0\textwidth]{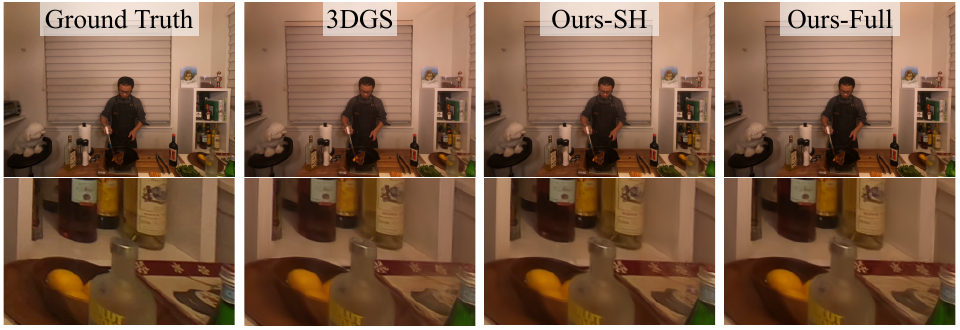}
   % \vspace{-7.5mm}
  \caption{\textbf{Qualitative comparison of 3DGS~\cite{3dg_2023}, Ours-SH and Ours-Full.} Conducted on the \textit{Flame Steak} scene from the Neural 3D Video Dataset~\cite{li2022neural}. 3DGS is trained per-frame. Ours-SH denotes replacing our feature rendering with spherical harmonics rendering in 3DGS~\cite{3dg_2023}.}
  \label{fig:img_comp}
   % \vspace{-3mm}
\end{figure*}

\subsection{Initialization of Features}
In our model, $\mathbf{f}^{base}$ and $\mathbf{f}^{dir}$ are initialized with the color of SfM points. $\mathbf{f}^{time}$ is initialized as zeros. 
The last three rows of~\cref{tab:changeNpNq} show an ablation of feature initialization, where our choice (Ours-Full) works better than random initialization.

\subsection{Feature Rendering vs. Spherical Harmonics}
In this experiment, we replace our full model's feature rendering with spherical harmonics rendering, and refer to this baseline as Ours-SH.
~\cref{tab:rendering} and~\cref{fig:img_comp} show that Ours-Full has better quantitative and visual quality while having smaller model size than Ours-SH. For fair comparisons of FPS, all methods use PyTorch implementation for rendering.

\begin{table}
\caption{\textbf{Comparison with per-frame 3DGS and replacing the MLP in Ours-Full with SH.} Conducted on the first 50 frames of \textit{Flame Salmon} and \textit{Flame Steak} scenes.
Ours-SH uses SH of order 0 to 3, as in 3DGS.
}
\label{tab:rendering}
\begin{center}
\resizebox{1\columnwidth}{!}{
\begin{tabular}{lccc|ccc}
\toprule
  & Size (MB)$\downarrow$ & FPS$\uparrow$  & Train Time (min.)$\downarrow$  & PSNR$\uparrow$ & DSSIM$_1$$\downarrow$ & LPIPS$\downarrow$ \\
\midrule
3DGS & 5100 & 135 & 700 & 29.76 & 0.0311 & 0.0486 \\
\midrule
Ours-SH & 118 & 132 & 37 & 31.37 & 0.0276 & 0.0470 \\
Ours-Full & \textbf{37}  & \textbf{145} & \textbf{35} & \textbf{31.66} & \textbf{0.0274} & \textbf{0.0467} \\
\bottomrule
\end{tabular}
}
\end{center}
% \vspace{-2mm}
\end{table}

\subsection{Comparison with Per-Frame 3DGS}
To validate the improvements of our method, we further compare with per-frame trained 3DGS.
As shown in~\cref{tab:rendering}, our method has much smaller size and better rendering quality.~\cref{fig:img_comp} shows visual comparison. 

\subsection{Longer Video Sequence}
In our experiments, following prior arts~\cite{song2023nerfplayer,attal2023hyperreel}, we train each model with 50-frame video sequence and arrange these models in series to render full-length sequences (typically 300 frames). In practice, this scheme can work for long videos at the cost of redundancy among models (\eg, static parts of the scene are repeatedly modeled). Our method also supports using a single model to represent more frames. 
Here, we conduct an experiment that directly trains our model with 300 frames on the \textit{Flame Salmon} scene from the Neural 3D Video Dataset~\cite{li2022neural}. 
As shown in~\cref{tab:n4dlong}, compared to six 50-frame models in series, our single 300-frame model can reduce the per-frame training time and model size by around $80\%$ and $30\%$ respectively.
At the same time, the rendering quality is comparable.
This is attributed to our temporal opacity formulation so that complex long-sequence motion can be represented by multiple simpler motion segments.

\begin{table*}
\caption{\label{tab:n4dlong}
    \textbf{Performance of longer sequence per model on the \textit{Flame Salmon} scene from the Neural 3D Video Dataset.} We increase the training frames per model from 50 to 300. Longer sequence per model has smaller model size and shorter per-frame training time. 
 }
\centering
 \resizebox{1.0\linewidth}{!}{%
\begin{tabular}{@{}l|c|c|cccc|c|c|r@{}}
\toprule
   Video Length per Model & \# of Models & Iterations per Model & PSNR$\uparrow$ & DSSIM$_1$$\downarrow$& DSSIM$_2$$\downarrow$ & LPIPS$\downarrow$ & Per-Frame Training Time (sec.)$\downarrow$ & Per-Frame Size (MB)$\downarrow$ & Total Size (MB)$\downarrow$ \\
\midrule

   50  frames  & 6  & 12K &\textbf{29.48} & 0.038  & 0.0224 & \textbf{0.063}  &20 & 1 & 300  \\

   300 frames & 1  & 10K & 29.17 & \textbf{0.037} & \textbf{0.0222} &  0.068 & \textbf{3.7}& \textbf{0.7} & \textbf{216}  \\
   
\bottomrule
\end{tabular}
 }
\end{table*}

\section{Discussions}
\label{sec::additionaldisscusion}

\begin{figure}
    \centering
    \includegraphics[width=1.0\linewidth]{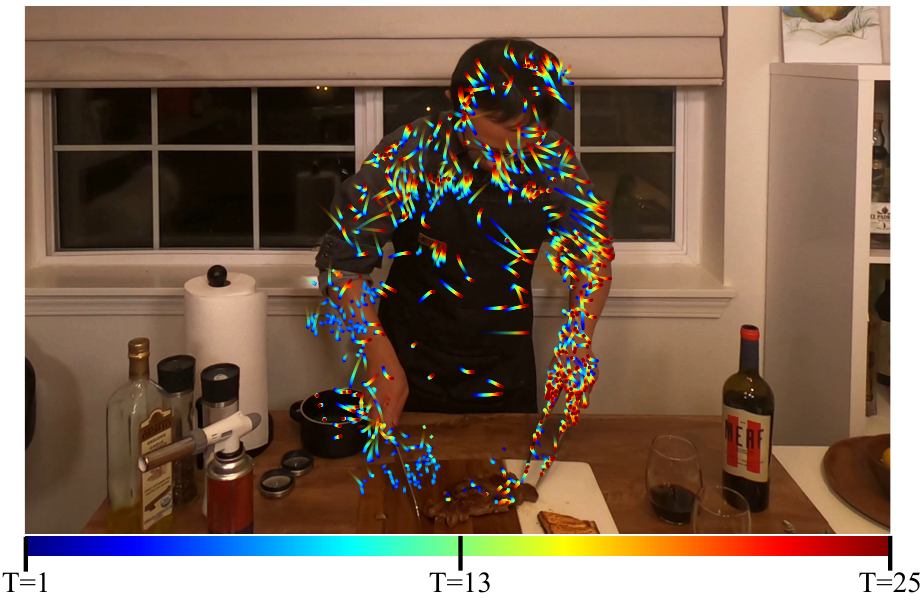}
    \caption{\textbf{Visualization of trajectory across 25 frames}. The background image is the ground truth at time $T=25$. The color of a trajectory denotes timestamp, where dark blue corresponds to $T=1$ and dark red corresponds to $T=25$. To visualize temporal opacity along a trajectory, we set the alpha channel value of a segment based on temporal opacity (excluding the spatial opacity term $\sigma_{i}^s$). We only show moving objects in the scene.}
    \label{fig:vistraj}
\end{figure}

Our method is able to model shadows and ambient occlusions, as demonstrated in~\cref{fig:comp_features} and~\cref{fig:img_comp}.
For complex motion, our temporal opacity allows using multiple Gaussians where each one only needs to fit a shorter and less complex motion segment.
Generally, the size of temporal RBF is small for fast-changing volumetric objects (\eg, flames) and large for static solid objects. 
Learned motion tends to be small for static objects and large for moving objects.~\cref{fig:vistraj} visualizes the temporal RBF and motion for an example scene.
Note that our method does not apply additional regularization on motion.

For guided sampling, although it can alleviate the blurring in areas that are insufficiently covered by sparse point cloud, it cannot fully eliminate such artifacts. This is reflected in some challenging scenarios such as the far content outside windows in the \textit{Coffee Martini} scene and the flame in the \textit{02\_Flames} scene. The reason is that we do not have accurate depth of these areas, hence need to spawn new Gaussians across a depth range. However, these Gaussians may not cover the exact correct locations, and the ones near the correct locations may also be pruned during subsequent training.
A possible solution would be to leverage the depth priors from learning-based depth estimation methods.

\section{Experiment Details}
\label{sec::expdetails}
\subsection{Baselines}
Since the open source code of MixVoxels~\cite{Wang2023ICCV} does not contain the training config for MixVoxels-X, we use MixVoxels-L in our comparisons.
We train HyperReel with their official code to generate visual examples. 
Note that the training time of HyperReel for each scene on the Technicolor Dataset is about 3.5 hours, while that of our full model on the Technicolor Dataset is only about 1 hour. 
For Dynamic 3DGS, its performance on the Neural 3D Video Dataset is not reported in their original paper.
When applying their open source code to Neural 3D Video Dataset with default hyperparameters, the rendering quality is subpar.
So we tune its hyperparameters to improve the performance on this dataset.

\subsection{Camera Models}
We use the original \textit{centered-undistorted} camera model from 3DGS~\cite{3dg_2023} for the Neural 3D Video Dataset.
We implement the \textit{uncentered-undistorted} camera model for the Technicolor Dataset. 
For the Google Immersive Dataset~\cite{broxton2020immersive}, to evaluate on the same distorted videos as~\cite{attal2020, song2023nerfplayer}, we further adapt our method to fit the \textit{uncentered-distorted} camera model with a differentiable image space warping, which maps perspective view to fish-eye distorted view. 
For the real-time demo, we retrain our models on the undistorted videos for simplicity. As there are black pixels in the undistorted images, we opt to use a mask to mask out the black pixels. Since image warping and masking are differentiable, our models can still be trained end-to-end. 

\subsection{Initialization}
Following 3DGS~\cite{3dg_2023}, we use the sparse point cloud from COLMAP for initialization. Since the datasets provide camera intrinsics and extrinsics, we input them to COLMAP and call \textit{point triangulator} to generate sparse points. The running time for \textit{point triangulator} is much less than that of dense reconstruction. For the \textit{Theater}, \textit{Train} and \textit{Birthday} scenes in~\cref{tab:quant_dynamic} of the main paper and in~\cref{tab:technicolor_perscene}, we only use 25 percent of SfM points from each frame (except the first frame whose SfM points are all used). The selection of SfM points is based on the distance between each point and its nearest neighbor. After sorting the distances, we keep the points with the longest distance to reduce redundancy.
In the ablation study on the number of frames whose SfM points are used (\cref{tab:techniabsize} in the main paper), we use all the points in each sampled frames. 

Features $\mathbf{f}^{base}$ and $\mathbf{f}^{dir}$ are initialized with the color of SfM points. $\mathbf{f}^{time}$ is initialized as zeros.

\subsection{Density Control}
During training, we conduct 12 times of cloning/splitting and over 50 times of pruning on the Technicolor dataset. Sparse points from multiple timestamps contain richer but more redundant information than sparse points from a single timestamp (or static points). Thus, after densification and guided sampling steps, we gradually prune Gaussians with small spatial opacity to keep the most representative Gaussians.

\subsection{Guided Sampling}
We uniformly sample from $s\times d$ to $7.5\times d$ with small random noise. $d$ is the max depth in a training view. $s$ is set as $0.7$ for most scenes.

\subsection{Others}
We apply \textit{sigmoid} function to get the final RGB color for Neural 3D dataset and \textit{clamp} function for the other two datasets in our full model. 
We use a linear form of the time variable and do not apply positional encoding on it.
We use Nvidia RTX 3090 when reporting our rendering speed in the comparisons with other methods, and use Nvidia RTX 4090 in our real-time demos.

\section{More Results}
\label{sec::supplevisual}
We provide per-scene quantitative comparisons on the Neural 3D Video Dataset~\cite{li2022neural} (\cref{tab:n3d_perscene}), Google Immersive Dataset~\cite{broxton2020immersive} (\cref{tab:immersive_perscene}) and Technicolor Dataset~\cite{techni} (\cref{tab:technicolor_perscene}).
Our method outperforms the other baselines on most scenes.
We also provide per-scene Gaussian numbers of trained 50-frame models in \cref{tab:perscene_number}.

\begin{figure*}[h]
  \includegraphics[width=1.0\textwidth]{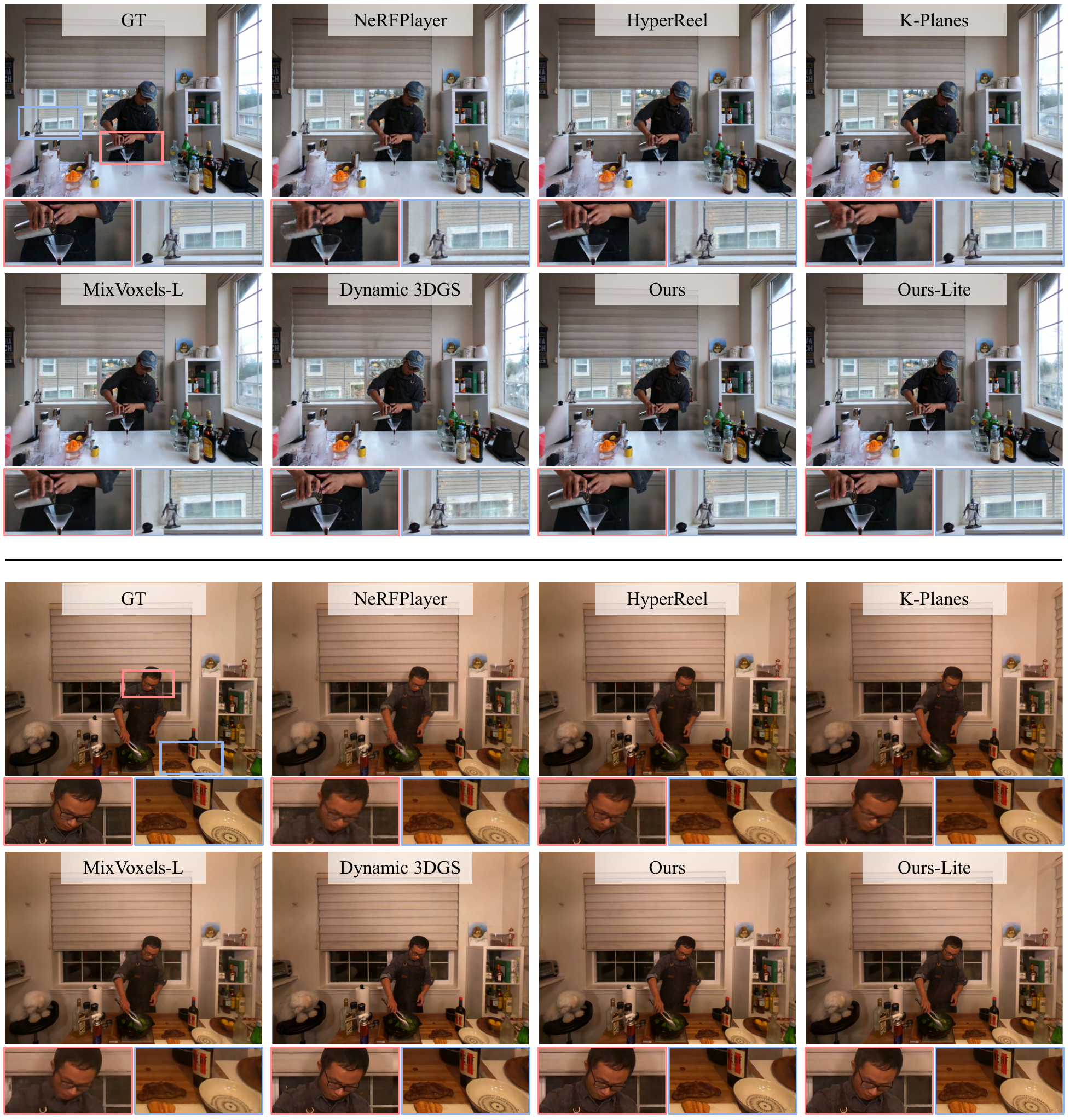}
  \caption{\textbf{Qualitative comparisons on the Neural 3D Video Dataset~\cite{li2022neural}.} To be continued in the next page.}
  \label{fig:comp_n4d_supp1}
\end{figure*}

\begin{figure*}[h]
  \includegraphics[width=1.0\textwidth]{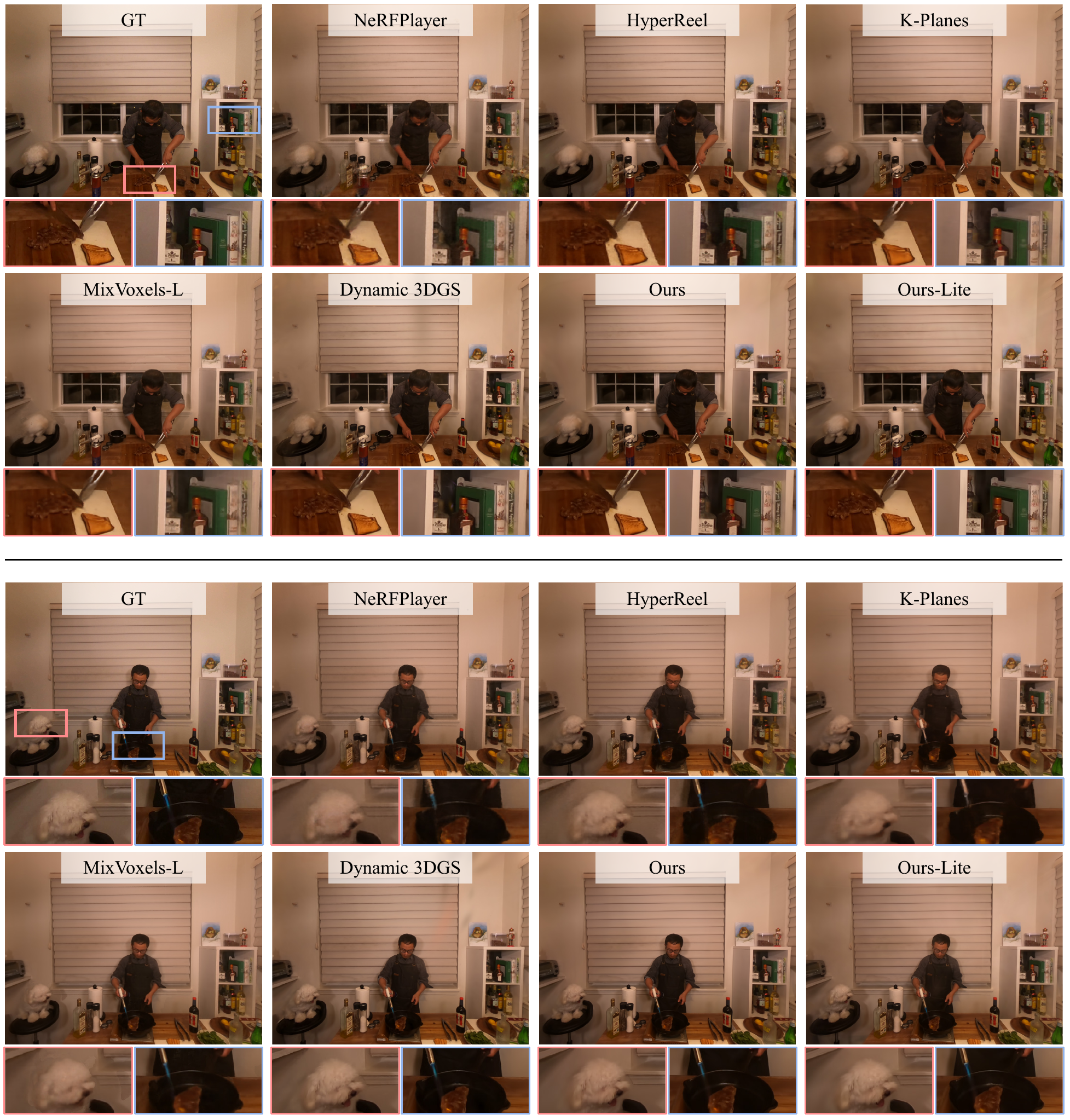}
  \caption{\textbf{Qualitative comparisons on the Neural 3D Video Dataset~\cite{li2022neural}.} To be continued in the next page.}
  \label{fig:comp_n4d_supp2}
\end{figure*}

\begin{figure*}[h]
  \includegraphics[width=1.0\textwidth]{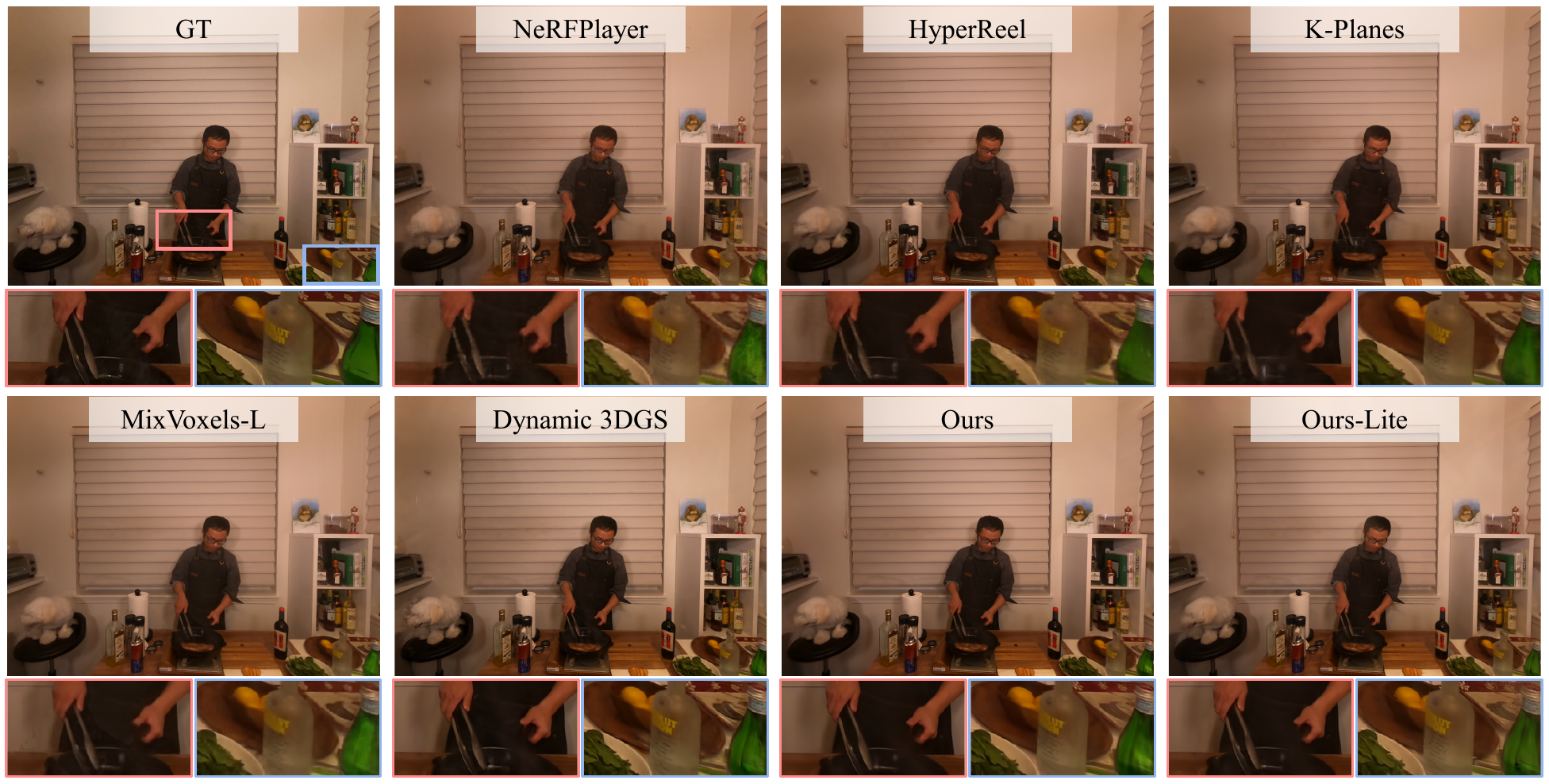}
  \caption{\textbf{Qualitative comparisons on the Neural 3D Video Dataset~\cite{li2022neural}.}}
  \label{fig:comp_n4d_supp3}
\end{figure*}

~\cref{fig:comp_n4d_supp1,fig:comp_n4d_supp2,fig:comp_n4d_supp3} show more visual comparisons of our full model and our lite-version model with NeRFPlayer~\cite{song2023nerfplayer}, HyperReel~\cite{attal2023hyperreel}, K-Planes~\cite{fridovich2023k}, MixVoxels-L~\cite{Wang2023ICCV} and Dynamic 3DGS~\cite{4DG3DV} on the Neural 3D Video Dataset~\cite{li2022neural}. 

\begin{figure*}[h]
  \includegraphics[width=1.0\textwidth]{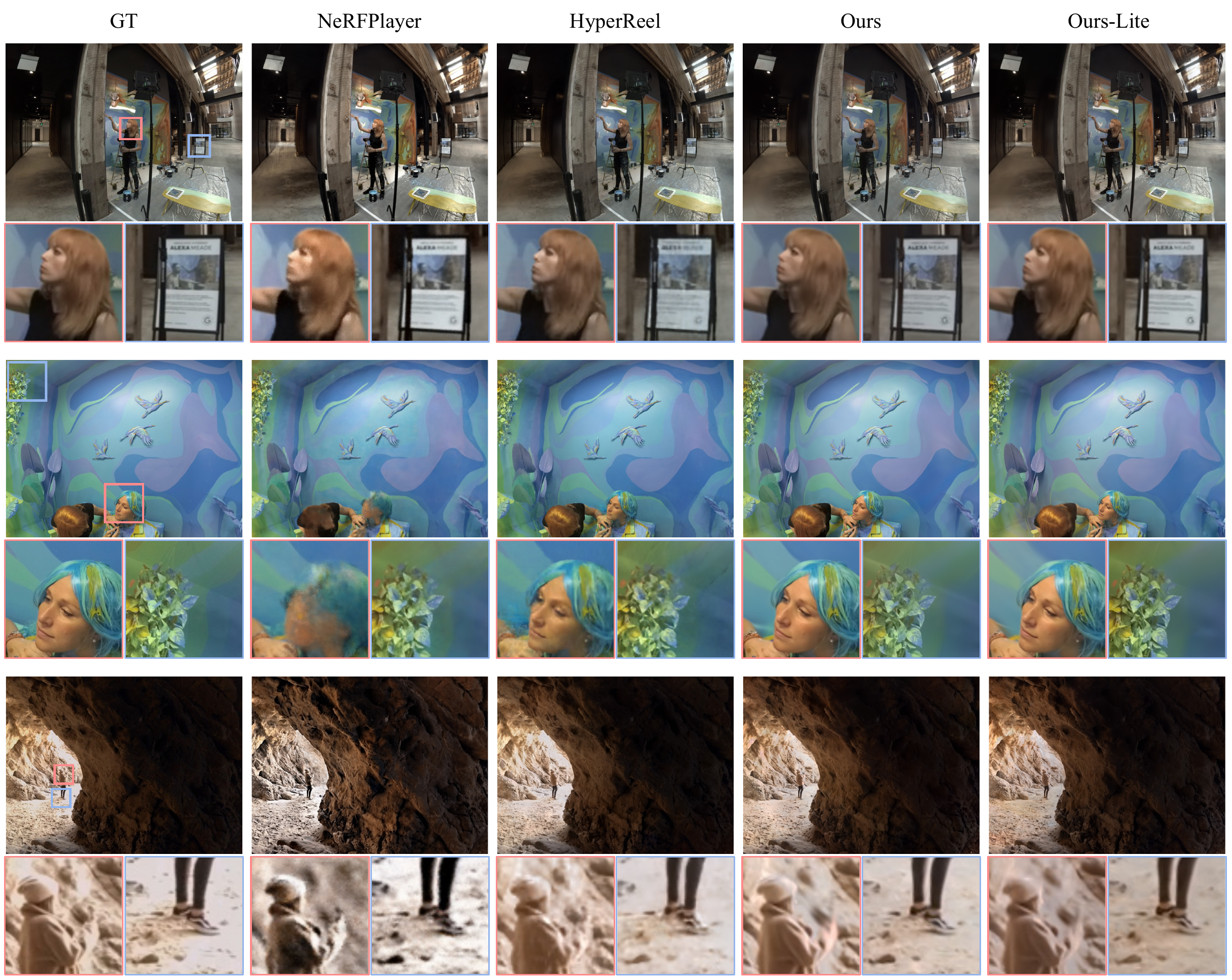}
  \caption{\textbf{Qualitative comparisons on the Google Immersive Dataset~\cite{broxton2020immersive}.}}
  \label{fig:comp_immersive_supp}
\end{figure*}

\cref{fig:comp_immersive_supp} shows more visual comparisons on the Google Immersive Dataset~\cite{broxton2020immersive}.
We compare the results of our full model and lite-version model to NeRFPlayer~\cite{song2023nerfplayer} and HyperReel~\cite{attal2023hyperreel}.

\begin{figure*}[h]
  \includegraphics[width=1.0\textwidth]{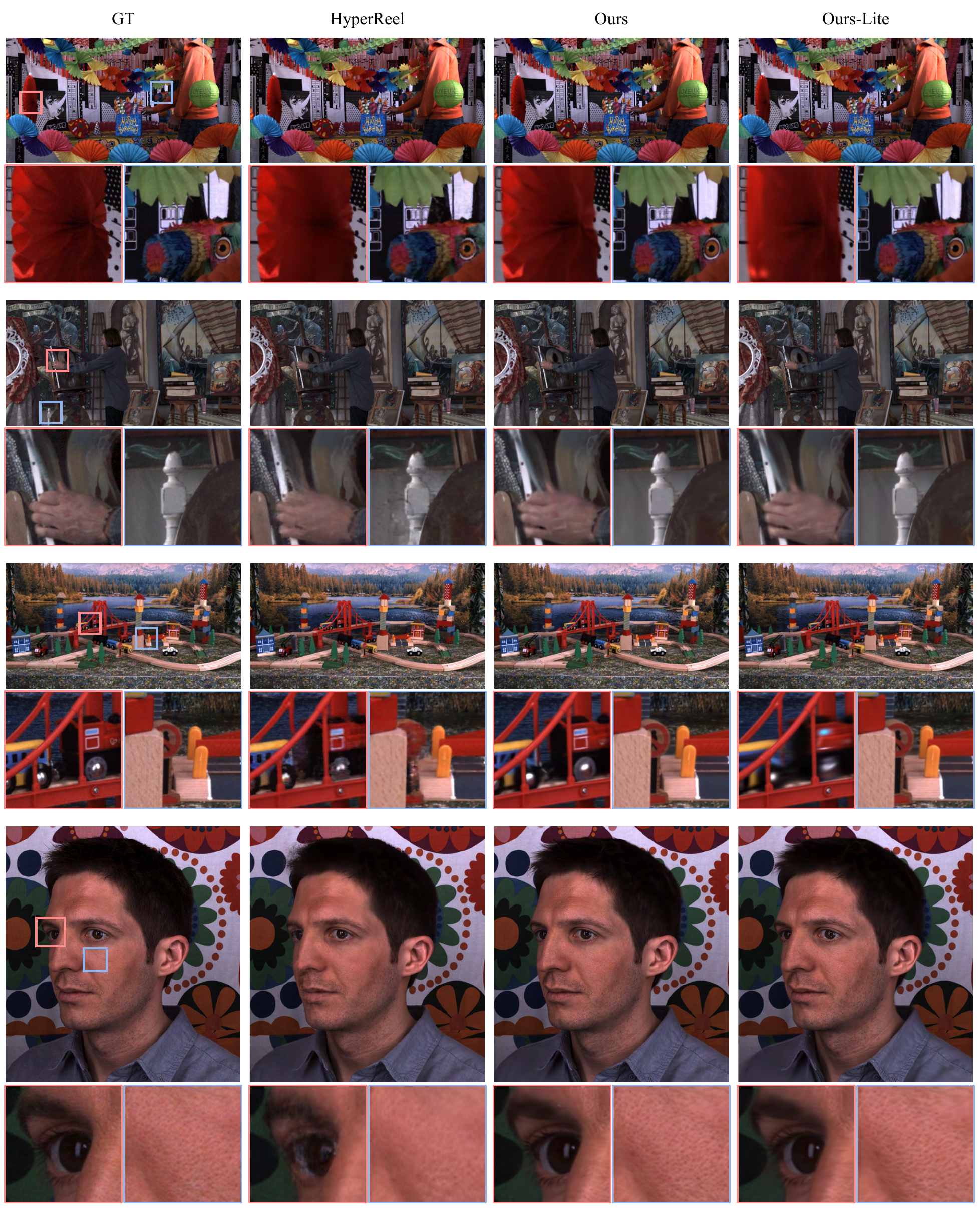}
  \caption{\textbf{Qualitative comparisons on the Technicolor Dataset~\cite{techni}.}}
  \label{fig:comp_technicolor_supp}
\end{figure*}

\cref{fig:comp_technicolor_supp} shows visual comparisons on the Technicolor Dataset~\cite{techni}.
We compare the results of our full model and lite-version model to HyperReel~\cite{attal2023hyperreel}.

The above visual comparisons demonstrate that our method preserves sharp details while containing fewer artifacts.
Compared to our full model, the results of our lite-version model are slightly blurrier.

\begin{table*}
\caption{
\textbf{Per-scene quantitative comparisons on the Neural 3D Video Dataset~\cite{li2022neural}.}
Some methods only report part of the scenes. 
$^1$ only includes the \textit{Flame Salmon} scene.
$^2$ excludes the \textit{Coffee Martini} scene.
``-" denotes results that are unavailable in prior work.
}
\label{tab:n3d_perscene}
\centering
\resizebox{1\linewidth}{!}{%
\begin{tabular}{l|l|llllll}
\toprule
Method & Avg. & \textit{Coffee Martini} & \textit{Cook Spinach} & \textit{Cut Roasted Beef} & \textit{Flame Salmon} & \textit{Flame Steak} & \textit{Sear Steak} \\
\midrule
\midrule
\multicolumn{2}{l}{\textbf{PSNR$\uparrow$}} \\
\midrule
Neural Volumes~\cite{Lombardi2019} $^1$ & $22.80$ & - & - & - & $22.80$ & - & - \\ 
LLFF~\cite{mildenhall2019llff} $^1$ & $23.24$ & - & - & - & $23.24$ & - & - \\ 
DyNeRF~\cite{li2022neural} $^1$ & $29.58$ & - & - & - & $29.58$ & - & - \\ 
HexPlane~\cite{cao2023hexplane} $^2$ & $31.71$ & - & $32.04$ & $32.55$ & $29.47$ & $32.08$ & $32.39$ \\
NeRFPlayer~\cite{song2023nerfplayer} & $30.69$ & $\textbf{31.53}$ & $30.56$ & $29.35$ & $31.65$ & $31.93$ & $29.13$ \\
HyperReel~\cite{attal2023hyperreel} & $31.10$ & $28.37$ & $32.30$ & $32.92$ & $28.26$ & $32.20$ & $32.57$ \\
K-Planes~\cite{fridovich2023k} & $31.63$ & $29.99$ & $32.60$ & $31.82$ & $30.44$ & $32.38$ & $32.52$ \\
MixVoxels-L~\cite{Wang2023ICCV} & $31.34$ & $29.63$ & $32.25$ & $32.40$ & $29.81$ & $31.83$ & $32.10$ \\
MixVoxels-X~\cite{Wang2023ICCV} & $31.73$ & $30.39$ & $32.31$ & $32.63$ & $\textbf{30.60}$ & $32.10$ & $32.33$ \\
Dynamic 3DGS~\cite{4DG3DV} & $30.67$ & $26.49$ & $32.97$ & $30.72$ & $26.92$ & $33.24$ & $33.68$ \\
\midrule
Ours & $\textbf{32.05}$ & $28.61$ & $\textbf{33.18}$ &$33.52$& $29.48$ & $\textbf{33.64}$ & $\textbf{33.89}$ \\
Ours-Lite  & $31.59$ & $27.49$ & $32.92$ & $\textbf{33.72}$ & $28.67$ & $33.28$ & $33.47$ \\
\midrule
\midrule

\multicolumn{2}{l}{\textbf{DSSIM$_1$$\downarrow$}} \\
\midrule
NeRFPlayer~\cite{song2023nerfplayer} & $0.034$ & $\textbf{0.0245}$ & $0.0355$ & $0.0460$ & $\textbf{0.0300}$ & $0.0250$ & $0.0460$ \\
HyperReel~\cite{attal2023hyperreel} & $0.036$ & $0.0540$ & $0.0295$ & $0.0275$ & $0.0590$ & $0.0255$ & $0.0240$ \\
Dynamic 3DGS~\cite{4DG3DV} & $0.035$ & $0.0557$ & $0.0263$ & $0.0295$ & $0.0512$ & $0.0233$ & $0.0224$ \\
\midrule
Ours & $\textbf{0.026}$ & $0.0415$ & $\textbf{0.0215}$ & $\textbf{0.0205}$ & $0.0375$ &$\textbf{0.0176}$ & $\textbf{0.0174}$ \\
Ours-Lite & $0.027$ & $0.0437$ & $0.0218$ & $0.0209$ & $0.0387$ & $0.0179$ &$0.0177$ \\
\midrule
\midrule

\multicolumn{2}{l}{\textbf{DSSIM$_2$$\downarrow$}} \\
\midrule
Neural Volumes~\cite{Lombardi2019} $^1$ & $0.062$ & - & - & - & $0.062$ & - & - \\ 
LLFF~\cite{mildenhall2019llff} $^1$ & $0.076$ & - & - & - & $0.076$ & - & - \\ 
DyNeRF~\cite{li2022neural} $^1$ & $0.020$ & - & - & - & $\textbf{0.020}$ & - & - \\ 
K-Planes~\cite{fridovich2023k} & $0.018$ & $0.0235$ & $0.0170$ & $0.0170$ & $0.0235$ & $0.0150$ & $0.0130$ \\
MixVoxels-L~\cite{Wang2023ICCV} & $0.017$ & $0.0244$ & $0.0162$ & $0.0157$ & $0.0255$ & $0.0144$ & $0.0122$ \\
MixVoxels-X~\cite{Wang2023ICCV} & $0.015$ & $\textbf{0.0232}$ & $0.0160$ & $0.0146$ & $0.0233$ & $0.0137$ & $0.0121$ \\
Dynamic 3DGS~\cite{4DG3DV} & $0.019$ & $0.0332$ & $0.0129$ & $0.0161$ & $0.0302$ & $0.0113$ & $0.0105$ \\
\midrule
Ours & $\textbf{0.014}$ & $0.0250$ & $\textbf{0.0113}$ & $\textbf{0.0105}$ & $0.0224$ & $\textbf{0.0087}$ & $\textbf{0.0085}$ \\
Ours-Lite & $0.015$ & $0.0270$ & $0.0118$ & $0.0112$ & $0.0244$ & $0.0097$ & $0.0095$ \\
\midrule
\midrule

\multicolumn{2}{l}{\textbf{LPIPS$_{Alex}$$\downarrow$}} \\
\midrule
Neural Volumes~\cite{Lombardi2019} $^1$ & $0.295$ & - & - & - & $0.295$ & - & - \\ 
LLFF~\cite{mildenhall2019llff} $^1$ & $0.235$ & - & - & - & $0.235$ & - & - \\ 
DyNeRF~\cite{li2022neural} $^1$ & $0.083$ & - & - & - & $0.083$ & - & - \\ 
HexPlane~\cite{cao2023hexplane} $^2$ & $0.075$ & - & $0.082$ & $0.080$ & $0.078$ & $0.066$ & $0.070$ \\
NeRFPlayer~\cite{song2023nerfplayer} & $0.111$ & $0.085$ & $0.113$ & $0.144$ & $0.098$ & $0.088$ & $0.138$ \\
HyperReel~\cite{attal2023hyperreel} & $0.096$ & $0.127$ & $0.089$ & $0.084$ & $0.136$ & $0.078$ & $0.077$ \\
MixVoxels-L~\cite{Wang2023ICCV} & $0.096$ & $0.106$ & $0.099$ & $0.088$ & $0.116$ & $0.088$ & $0.080$ \\
MixVoxels-X~\cite{Wang2023ICCV} & $0.064$ & $0.081$ & $0.062$ & $0.057$ & $0.078$ & $0.051$ & $0.053$ \\
Dynamic 3DGS~\cite{4DG3DV} & $0.099$ & $0.139$ & $0.087$ & $0.090$ & $0.122$ & $0.079$ & $0.079$ \\
\midrule
Ours& $\textbf{0.044}$ & $\textbf{0.069}$ & $\textbf{0.037}$ & $\textbf{0.036}$ & $\textbf{0.063}$ & $\textbf{0.029}$ & $\textbf{0.030}$ \\
Ours-Lite & $0.047$ & $0.075$ & $0.038$ & $0.038$ & $0.068$ & $0.031$ & $0.031$ \\
\bottomrule
\end{tabular}
}
\end{table*}

\begin{table*}
\caption{
\textbf{Per-scene quantitative comparisons on the Google Immersive Dataset~\cite{broxton2020immersive}.}
}
\label{tab:immersive_perscene}
\centering
\resizebox{1\linewidth}{!}{
\begin{tabular}{l|c|ccccccc}
\toprule
Method & Avg. & \textit{01\_Welder} & \textit{02\_Flames} & \textit{04\_Truck} & \textit{09\_Exhibit} & \textit{10\_Face\_Paint\_1} & \textit{11\_Face\_Paint\_2} & \textit{12\_Cave} \\
\midrule
\midrule
\multicolumn{2}{l}{\textbf{PSNR$\uparrow$}} \\
\midrule
NeRFPlayer~\cite{song2023nerfplayer} & $25.8$ & $25.568$ & $26.554$ & $27.021$ & $24.549$ & $27.772$ & $27.352$ & $21.825$ \\
HyperReel~\cite{attal2023hyperreel} & $28.8$ & $25.554$ & $\textbf{30.631}$ & $27.175$ & $\textbf{31.259}$ & $29.305$ & $27.336$ & $\textbf{30.063}$ \\

\midrule
Ours& $\textbf{29.2}$ & $\textbf{26.844}$ & $30.566$ & $\textbf{27.308}$ & $29.336$ & $\textbf{30.588}$ & $\textbf{29.895}$ & $29.610$ \\
Ours-Lite & $27.5$ & $25.499$ & $29.505$ & $24.204$ & $27.973$ & $28.646
$ & $28.456$ & $27.977$ \\
\midrule
\midrule

\multicolumn{2}{l}{\textbf{DSSIM$_1$$\downarrow$}} \\
\midrule
NeRFPlayer~\cite{song2023nerfplayer} & $0.076$ & $0.0910$ & $0.0790$ & $0.0615$ & $0.0655$ & $0.0420$ & $0.0490$ & $0.1425$ \\
HyperReel~\cite{attal2023hyperreel} & $0.063$ & $0.1050$ & $0.0475$ & $0.0760$ & $0.0485$ & $0.0435$ & $0.0605$ & $0.0595$ \\

\midrule
Ours& $\textbf{0.042}$ & $\textbf{0.0504}$ & $\textbf{0.0349}$ & $\textbf{0.0524}$ & $\textbf{0.0447}$ & $\textbf{0.0240}$ & $\textbf{0.0320}$ & $\textbf{0.0543}$ \\
Ours-Lite& $0.051$ & $0.0585$ & $0.0546$ & $0.0684$ & $0.0516$ & $0.0271$ & $0.0326$ & $0.0630$ \\
\midrule
\midrule

\multicolumn{2}{l}{\textbf{LPIPS$_{Alex}$$\downarrow$}} \\
\midrule
NeRFPlayer~\cite{song2023nerfplayer} & $0.196$ & $0.289$ & $0.154$ & $0.164$ & $0.151$ & $0.147$ & $0.152$ & $0.314$ \\
HyperReel~\cite{attal2023hyperreel} & $0.193$ & $0.281$ & $0.159$ & $0.223$ & $0.140$ & $0.139$ & $0.195$ & $0.214$ \\

\midrule
Ours & $\textbf{0.081}$ & $\textbf{0.098}$ & $\textbf{0.059}$ & $\textbf{0.087}$ & $\textbf{0.073}$ & $\textbf{0.055}$ & $0.063$ & $\textbf{0.133}$ \\
Ours-Lite & $0.095$ & $0.119$ & $0.070$ & $0.115$ & $0.087$ & $0.067$ & $\textbf{0.062}$ & $0.143$ \\
\bottomrule
\end{tabular}
}
\end{table*}

\begin{table*}
\caption{
\textbf{Per-scene quantitative comparisons on the Technicolor Dataset~\cite{techni}.}
}
\label{tab:technicolor_perscene}
\centering
% \resizebox{1\linewidth}{!}{
\begin{tabular}{l|c|ccccc}
\toprule
Method & Avg. & \textit{Birthday} & \textit{Fabien} & \textit{Painter} & \textit{Theater} & \textit{Trains} \\
\midrule
\midrule
\multicolumn{2}{l}{\textbf{PSNR$\uparrow$}} \\
\midrule
DyNeRF~\cite{li2022neural} & $31.8$ & $29.20$ & $32.76$ & $35.95$ & $29.53$ & $31.58$ \\
HyperReel~\cite{attal2023hyperreel} & $32.7$ & $29.99$ & $34.70$ & $35.91$ & $\textbf{33.32}$ & $29.74$ \\

\midrule
Ours& $\textbf{33.6}$ & $\textbf{32.09}$ & $\textbf{35.70}$ & $\textbf{36.44}$ & $30.99$ & $\textbf{32.58}$ \\
Ours-Lite & $33.0$ & $31.59$ & $35.28$ & $35.95$ & $30.12$ & $32.17$ \\
\midrule
\midrule

\multicolumn{2}{l}{\textbf{DSSIM$_1$$\downarrow$}} \\
\midrule
HyperReel~\cite{attal2023hyperreel} & $0.047$ & $0.0390$ & $0.0525$ & $0.0385$ & $\textbf{0.0525}$ & $0.0525$ \\

\midrule
Ours& $\textbf{0.040}$ & $\textbf{0.0290}$ & $\textbf{0.0471}$ & $\textbf{0.0366}$ & $0.0596$ & $\textbf{0.0294}$ \\
Ours-Lite & $0.044$ & $0.0330$ & $0.0522$ & $0.0382$ & $0.0634$ & $0.0324$ \\
\midrule
\midrule

\multicolumn{2}{l}{\textbf{DSSIM$_2$$\downarrow$}} \\
\midrule
DyNeRF~\cite{li2022neural} & $0.021$ & $0.0240$ & $\textbf{0.0175}$ & $\textbf{0.0140}$ & $0.0305$ & $0.0190$ \\

\midrule
Ours & $\textbf{0.019}$ & $\textbf{0.0153}$ & $0.0179$ & $0.0146$ & $\textbf{0.0287}$ & $\textbf{0.0168}$ \\
Ours-Lite & $0.021$ & $0.0175$ & $0.0201$ & $0.0154$ & $0.0312$ & $0.0185$ \\
\midrule
\midrule

\multicolumn{2}{l}{\textbf{LPIPS$_{Alex}$$\downarrow$}} \\
\midrule
DyNeRF~\cite{li2022neural} & $0.140$ & $0.0668$ & $0.2417$ & $0.1464$ & $0.1881$ & $0.0670$ \\
HyperReel~\cite{attal2023hyperreel} & $0.109$ & $0.0531$ & $0.1864$ & $0.1173$ & $\textbf{0.1154}$ & $0.0723$ \\

\midrule
Ours & $\textbf{0.084}$ & $\textbf{0.0419}$ & $\textbf{0.1141}$ & $\textbf{0.0958}$ & $0.1327$ & $\textbf{0.0372}$ \\
Ours-Lite & $0.097$ & $0.0532$ & $0.1359$ & $0.0989$ & $0.1487$ & $0.0492$ \\
\bottomrule
\end{tabular}
% }
\end{table*}

\begin{table*}
\caption{
\textbf{Per-scene Gaussian numbers (K) on three datasets.}
For each scene, the number is averaged over 50-frame models.
}
\label{tab:perscene_number}
\centering
\resizebox{1.0\linewidth}{!}{
\begin{tabular}{l|c|ccccccc}
\toprule
 & Avg. & \textit{01\_Welder} & \textit{02\_Flames} & \textit{04\_Truck} & \textit{09\_Exhibit} & \textit{10\_Face\_Paint\_1} & \textit{11\_Face\_Paint\_2} & \textit{12\_Cave} \\
\midrule
Google Immersive Dataset~\cite{broxton2020immersive} & 427 & 571 & 389 & 374 & 484 & 285 & 249 & 629 \\
\midrule
\midrule
 & Avg. &  \textit{Coffee Martini} & \textit{Cook Spinach} & \textit{Cut Roasted Beef} & \textit{Flame Salmon} & \textit{Flame Steak} & \textit{Sear Steak}   \\
\midrule
Neural 3D Video Dataset~\cite{li2022neural} & 215 & 262
 & 189 & 169
 & 319 & 177 & 176  \\
\midrule
\midrule
 & Avg. & \textit{Birthday} &  \textit{Fabien} & \textit{Painter} & \textit{Theater} & \textit{Trains}  \\
\midrule
Technicolor Dataset~\cite{techni} & 374 & 379 & 295 & 412 & 313 & 470
 \\

\bottomrule
\end{tabular}
}
\end{table*}

\end{document}